\documentclass[twoside]{article}

\usepackage[accepted]{aistats2025}
\usepackage{array}
\usepackage{textcomp}
\usepackage{stfloats}
\usepackage{url}
\usepackage{verbatim}
\usepackage{graphicx}

\usepackage{amssymb} 
\usepackage{multirow}
\usepackage{comment}
\usepackage[marginal]{footmisc}
\usepackage{amsfonts}
\usepackage[ruled]{algorithm2e}
\usepackage{algorithmic,algorithm2e}
\usepackage{mathtools}
\usepackage{booktabs}
\usepackage{color}
\usepackage{floatrow}
\usepackage{marvosym}
\usepackage{ntheorem}
\usepackage{url}
\usepackage{verbatim}
\usepackage{hyperref}
\usepackage{cleveref}
\usepackage{todonotes}
\usepackage{marginnote}
\usepackage{threeparttable}
\usepackage{svg}
\usepackage{subcaption}
\usepackage{natbib}

\usepackage{floatrow}

\usepackage{amsmath,amsfonts,bm}

\newfloatcommand{capbtabbox}{table}[][\FBwidth]
\usepackage{blindtext}

\usepackage{xcolor}

\SetKwComment{Comment}{/* }{ */}
\SetKwInput{KwData}{Input}
\SetKwInput{KwResult}{Output}

\newcommand{\ky}[1]{\textcolor{teal}{#1}}

\newcommand{\R}{\ensuremath{\mathbb{R}}}

\def\x{\mathbf{x}}

\def\W{\mathbf{W}}
\def\uW{\mathbb{W}}

\def\y{\mathbf{y}}

\def\E{\mathcal{E}}

\def\G{\mathcal{G}}

\def\A{\mathcal{A}}

\def\B{\mathbf{B}}

\def\y{\mathbf{y}}

\def\N{\mathcal{N}}

\def\K{\mathcal{K}}
\def\Frm{\mathrm{F}}

\DeclareMathOperator{\vct}{vec}

\newtheorem{theorem}{Theorem}[section]

\newtheorem{lemma}[theorem]{Lemma}
\newtheorem{propensity}[theorem]{Propensity}

\def\eqref#1{equation~\ref{#1}}

\def\1{\bm{1}}

\def\va{{\bm{a}}}
\def\vb{{\bm{b}}}

\def\ve{{\bm{e}}}

\def\vw{{\bm{w}}}
\def\vx{{\bm{x}}}
\def\vy{{\bm{y}}}
\def\vz{{\bm{z}}}

\def\mB{{\bm{B}}}

\def\mD{{\bm{D}}}
\def\mE{{\bm{E}}}

\def\mI{{\bm{I}}}

\def\mL{{\bm{L}}}
\def\mM{{\bm{M}}}

\def\mP{{\bm{P}}}
\def\mQ{{\bm{Q}}}

\def\mT{{\bm{T}}}
\def\mU{{\bm{U}}}
\def\mV{{\bm{V}}}
\def\mW{{\bm{W}}}
\def\mX{{\bm{X}}}
\def\mY{{\bm{Y}}}

\DeclareMathAlphabet{\mathsfit}{\encodingdefault}{\sfdefault}{m}{sl}
\SetMathAlphabet{\mathsfit}{bold}{\encodingdefault}{\sfdefault}{bx}{n}
\newcommand{\tens}[1]{\bm{\mathsfit{#1}}}

\def\tW{{\tens{W}}}

\def\gA{{\mathcal{A}}}

\def\gE{{\mathcal{E}}}

\def\gG{{\mathcal{G}}}

\def\gN{{\mathcal{N}}}
\def\gO{{\mathcal{O}}}
\def\gP{{\mathcal{P}}}

\def\gR{{\mathcal{R}}}

\def\gV{{\mathcal{V}}}

\def\sW{{\mathbb{W}}}

\DeclareMathOperator*{\argmax}{arg\,max}
\DeclareMathOperator*{\argmin}{arg\,min}

\begin{document}

%

%

\twocolumn[

\runningtitle{Heterogeneous Graph Structure Learning through the Lens of Data-generating Processes}

\aistatstitle{Heterogeneous Graph Structure Learning\\ through the Lens of Data-generating Processes}

\runningauthor{Keyue Jiang, Bohan Tang, Xiaowen Dong, Laura Toni}

\aistatsauthor{ Keyue Jiang$^1$ \And Bohan Tang$^{2}$ \And  Xiaowen Dong$^{2}$ \And Laura Toni$^1$ }

\aistatsaddress{ $^1$University College London, London, UK \And  $^{2}$University of Oxford, Oxford, UK } ]

\begin{abstract}
Inferring the graph structure from observed data is a key task in graph machine learning to capture the intrinsic relationship between data entities. While significant advancements have been made in learning the structure of homogeneous graphs, many real-world graphs exhibit heterogeneous patterns where nodes and edges have multiple types. This paper fills this gap by introducing the first approach for heterogeneous graph structure learning (HGSL). To this end, we first propose a novel statistical model for the data-generating process (DGP) of heterogeneous graph data, namely hidden Markov networks for heterogeneous graphs (H2MN). Then we formalize HGSL as a maximum a-posterior estimation problem parameterized by such DGP and derive an alternating optimization method to obtain a solution together with a theoretical justification of the optimization conditions. Finally, we conduct extensive experiments on both synthetic and real-world datasets to demonstrate that our proposed method excels in learning structure on heterogeneous graphs in terms of edge type identification and edge weight recovery.

\end{abstract}

\section{INTRODUCTION}
Graphs are a powerful and ubiquitous representation of relational data for real applications such as social networks~\citep{DBLP:journals/tsipn/SalamiYS17}, e-commerce~\citep{DBLP:conf/sigir/Wang0WFC19} and financial transactions~\citep{liu2018heterogeneous}, in addition to various scientific and technological areas.
However, a meaningful graph is not always readily available from the data~\citep{DBLP:journals/tsp/DongTFV16}, and sometimes the graph observed is not always clean~\citep{DBLP:journals/tsp/ChenSMK15}. Learning or inferring the underlying graph structure from observed data living in the nodes is, therefore, an important problem in the field of both graph machine learning ~\citep{DBLP:journals/jmlr/ChamiAPR022, DBLP:journals/debu/HamiltonYL17} and graph signal processing ~\citep{DBLP:journals/spm/DongTRF19}, ~\citep{DBLP:journals/spm/DongTTBF20, 8347162}. In the former, e.g., graph neural networks, graph structure learning (GSL) is typically plugged in as an extra component to refine the graph topology, and empower the prediction ability of models~\citep{DBLP:journals/mia/ZaripovaCKABN23}, ~\citep{DBLP:journals/corr/abs-2305-16174}. In the latter, GSL assists in downstream algorithms, such as spectral clustering~\citep{DBLP:journals/tsp/DongTFV16}, cooperation game~\citep{DBLP:conf/icml/RossiMLB022}, 
video prediction~\citep{zhong2024motion}, etc.  

GSL algorithms~\citep{DBLP:journals/tsp/DongTFV16, DBLP:conf/aistats/Kalofolias16, DBLP:journals/tsipn/PuCDS21} assume that the features/signals on the graphs admit certain regularity or smoothness modeled by an underlying data-generating process (DGP). The common choices of the DGP for graph-structured data are Ising models~\citep{Ising1925BeitragZT}, Gaussian graphical models~\citep{yuan2007model, DBLP:journals/simods/JiaB22}, and pair-wise exponential Markov random fields (PE-MRF)~\citep{DBLP:conf/aistats/ParkHBL17}, where the features are assumed to be emitted from a multi-variate Gaussian distribution whose covariance matrix is uniquely determined by the graph structure. In such a scenario, the GSL problem is solved by the precision matrix estimation~\citep{yuan2007model, friedman2008sparse, DBLP:journals/jmlr/BanerjeeGd08} from the graph features. Following this framework, various algorithms have proposed to inject structural prior~\citep{pu2021learning}, guarantee graph connectivity~\citep{DBLP:journals/tsp/DongTFV16, pmlr-v51-kalofolias16}, or ensure convergence through relaxed optimization ~\citep{DBLP:journals/jstsp/EgilmezPO17}. 

The aforementioned methods are limited to homogeneous GSL tasks where nodes and edges in the graph belong to a single type, preventing the extension to heterogeneous graphs. Real-world graphs often exhibit heterogeneous patterns~\citep{DBLP:journals/corr/abs-2011-14867}, with multiple types of nodes and edges representing different kinds of entities and relationships. For instance, in a network representing a recommender system ~\citep{DBLP:conf/www/0004ZMK20}, nodes can have distinct types (e.g. users and items), and edges can represent different types of relations (e.g. like/dislike for user-item edges or following/being followed for user-user edges). Other examples include social networks, academic networks~\citep{DBLP:conf/kdd/LvDLCFHZJDT21, DBLP:conf/nips/GaoLFSH09}, and knowledge graphs~\citep{bollacker2008freebase, dettmers2018convolutional}. Effectively inferring the structure of heterogeneous graphs from node observations is an important challenge as for unsupervised discovery of complex relations within these social systems, but one that remains under-explored in the graph learning literature.

In this work, we aim to address this challenge and solve the problem of \emph{heterogeneous} graph structure learning (HGSL), a new formulation that applies to a wide range of graphs, including bipartite, multi-relational, and knowledge graphs. 
Our contributions can be summarized as follows. 

$\bullet$ \textbf{Problem formulation.} We formulate the HGSL problem as a maximum a-posterior (MAP) estimation of the adjacency tensor that captures the structure of heterogeneous graphs with different node/edge types. 

$\bullet$ \textbf{Novel data-generating process.} 
Solving the HGSL problem requires a proper design of the DGP for heterogeneous graphs. We develop a novel DGP based on hidden Markov networks~\citep{ghahramani2001introduction} for the heterogeneous graph data, namely hidden Markov networks for heterogeneous graphs (H2MN). 
    
$\bullet$ \textbf{Algorithm design.} 
Jointly estimating model parameters and graph structure is challenging as it is over-parameterized and leads to a non-convex optimization problem. 
Thus, we develop an effective algorithm to solve the HGSL problem and provide a theoretical justification for the optimization conditions. 

$\bullet$ \textbf{Empirical study.} Extensive experiments on both synthetic and real-world datasets were conducted to test the efficacy of our algorithm. The experimental results demonstrate that our proposed method consistently excels in structure learning tasks in terms of both edge type identification and edge weight recovery. 
Furthermore, we give illustrative examples of learned heterogeneous graphs in real-world network systems.

\section{PRELIMINARIES AND PROBLEM FORMULATION}


\subsection{Heterogeneous Graphs}
We now introduce the main notation needed in the paper to formalize the HGSL problem. 
Let $\G=\{\gV, \E\}$ denote an undirected graph with edges $\E$ and nodes $\gV$. In the context of the heterogeneous graph (HG), each node $v\in \gV$ is characterized by a node type $\phi(v) \in \A$ and each edge is assigned with a relation type $r \in \mathcal{R}$, where $\mathcal{A}$ and $\mathcal{R}$ are predefined node and edge type sets. 
Each edge in the HG is represented by a triplet $\{v, u, r\}$, which means nodes $v$ and $u$ are connected by relation type $r$. This formalizes a 3-D weighted tensor, $\tW = \{w_{vur}\}\in\mathbb{R}^{|\mathcal{V}|\times |\mathcal{V}| \times |\mathcal{R}|}$, with $|\cdot|$  the set cardinality.
For each node pair, $\tW_{vu:}$ is a length-$|\mathcal{R}|$ vector with only one nonzero entry indicating the weight and type for the edge connecting $u$ and $v$. We assume that the type of edge is determined by the types of connected nodes, i.e., if node $u$ is type `actor' and $v$ is `movie', the relation type $r$ can only be in the subset of $\{$`star in', `support in'$\}$, denoted as $\mathcal{R}_{\phi(v), \phi(u)}$. This is a conventional assumption in the literature~\citep{DBLP:conf/kdd/LvDLCFHZJDT21, DBLP:conf/www/0004ZMK20, DBLP:conf/www/GuoDBFMCH0Z23, DBLP:conf/nips/HuFZDRLCL20, DBLP:conf/iclr/ZhangC20}  and it reduces the degree of freedom in determining relation types.

Furthermore, each node $v$ is associated with a series of \textit{node features} that can either be \textit{signals} or \textit{labels} for the classes. The observable features are different depending on the datasets. For example, in the movie recommendation datasets~\citep{DBLP:journals/tiis/HarperK16}, the node features only contain labels for movie genres. In the city traffic network~\citep{datasets_urban, urban_datasets2}, no node label is observed while the traffic volume is viewed as the signal.  
 
\textbf{The node signals.} The node signals can be represented by a function $f: \gV \rightarrow \mathbb{R}^K$, which assigns a vector $\vx_v \in \mathbb{R}^{K}$ to node $v\in\mathcal{V}$.  Some datasets~\citep{DBLP:conf/kdd/LvDLCFHZJDT21, DBLP:journals/corr/abs-1102-2166, DBLP:conf/www/GuoDBFMCH0Z23} have a type-specific dimension for each node type. This study will focus on the cases with unified dimension sizes but the proposed method can be generalized to any heterogeneous graph by projecting the signals into a universal $\mathbb{R}^K$ with extra linear modules. Note that we do not consider edge signals encoding edge attributes and leave it for future work.

\textbf{The node labels. }
Within each node type, a node could contain a class label. For instance, node $v$ can have types ``actor'' and ``movie'' in a movie review graph, denoted by $\phi(v)$. Within each type,  nodes are classified into genres ``adventure'' or ``action'', denoted by $\vy_v$. The label is represented by a one-hot vector $\vy_v \in \R^{C_{\phi(v)}}$ 
where $C_{\phi(v)}$ is the number of classes. 

\textbf{The relation-wise connectivity matrix.} For each $r$, we define a probability matrix $\mB_r \in \R^{|C_{\phi(v)}| \times |C_{\phi(u)}|}$, where each entry of the matrix encodes the probability of two nodes belonging to specific classes to be connected with $r$. Conceptually, $\mB_{r}[p,q]=P(w_{uvr}=1\mid \vy_{u,p} = \vy_{v, q}=1)$ with $\vy_{u,p}$ the p-th entry of the vector $\vy_u$. Following the example of the movie review dataset, actors and movies in the same genre are more likely (with high probability) to be connected by a ``star in'' edge type. To ensure $\mB_r$ formalizes a valid probability matrix, the sum of the entries is, $\|\mB_r\|_1 =1$.

\subsection{Problem Formulation}
\label{subsec: HGSL_problem_for}

This paper focuses on learning heterogeneous graph structure from features living on nodes, together with the optimal parameters as a bi-product. Denoting the extra parameters for modeling the DGP as $\Theta$, we can formulate it as follows: 
Given nodes $\{v\}_{v \in \gV}$ with corresponding type $\{\phi(v)\}_{v \in \gV}$ and associated signals\footnote{To be concise, we first focus on the case where only the node signals $\vx_v$ are observable and leave the generalization to other scenarios in ~\cref{apdx: HGSL_labels}.} $\{\vx_v\}_{v\in \gV}$, together with a potential relation type set $\mathcal{R}$, we aim to learn a weighted and undirected heterogeneous graph $\G$ represented by tensor $\tW$ that encodes the weights and types of edges. 
Mathematically, this can be viewed as a MAP estimation problem, 
\begin{equation}
\label{eq: hetero_log_likelihood}
\begin{aligned}
\tW^*, &\Theta^* = \argmax_{\tW, \Theta} \log P(\tW, \Theta \mid  \{\vx_v\})\\
=&\argmin_{\tW, \Theta } -\log P(\{\vx_v\} \mid \tW, \Theta  ) + \Omega(\tW) +\Omega(\Theta),
\end{aligned}
\end{equation}
where $\Omega(\tW)$ and $\Omega(\Theta)$ are the negative-log priors.

\section{CONNECTING GENERATING PROCESS AND GRAPH STRUCTURE LEARNING}
\label{sec: DGP_GSL_connection}

\subsection{Hidden Markov Networks of Graph Features}
\label{subsec:prob_form}

\begin{figure}
\includegraphics[scale=0.4]{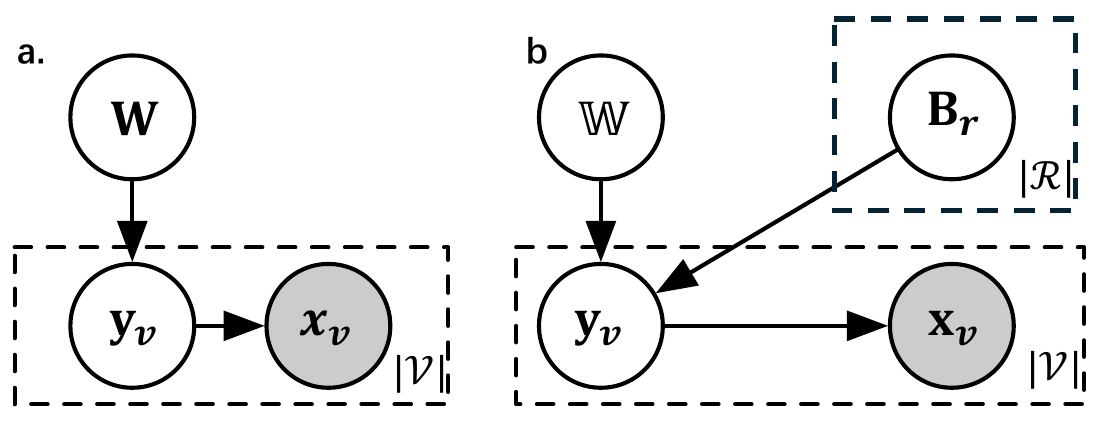}
 \caption{The graphical models for a) HMN and b) our H2MN. The shadowed variable is observable.}
 \label{fig: graphical}
\end{figure}

In this section, we introduce the usage of DGP, specifically hidden Markov networks (HMN)~\citep{ghahramani2001introduction}, to model homogeneous graphs, where $|\gR| = 1$ and $|\gA|=1$ so that $\sW$ reduces to a matrix $\mW = \{w_{uv}\}_{u,v\in\{1:|\gV|\}} \in \R^{|\gV| \times |\gV|}_{+}$ and $\mY$ has a unified dimension $C$. The graph Laplacian matrix is further defined as $\mL = \mD-\mW \in \mathbb{R}^{|\gV| \times |\gV|}$, with $\mD= \operatorname{diag}(\mW \mathbf{1})$ being the degree matrix and $d_v \equiv \mD_{vv}$. 
It is possible to stack the feature vector on each node together and obtain 
$\mX=\left[\vx_1, \vx_2, \ldots, \vx_{|\gV|}\right]^T \in \mathbb{R}^{{|\gV|}\times K}$ and $\mY=\left[\vy_1, \vy_2, \ldots, \vy_{|\gV|}\right]^T \in \mathbb{R}^{{|\gV|}\times C}$. 

A DGP on a graph models the underlying mechanism that generates data associated with the nodes and edges. Mathematically this is described by the joint probability density distribution (PDF) $P(\mX, \mY, \mW)$. Under the assumption of HMN, the hidden variable $\mY$\footnote{Conceptually, the labels are different from the hidden variables. In the derivation, we simplify the concept by drawing the equivalence similar to~\cite{DBLP:conf/nips/WeiYJB022} and view the hidden variables as the continuous embedding of labels. } is generated from a multivariate Gaussian distribution whose covariance matrix is parameterized by the graph structure $\mW$ and the signals $\mX$ are emitted from the hidden variables through an emission process $P(\mX \mid \mY)$. This yields a graphical model as in~\cref{fig: graphical}a and the joint PDF for the nodes and edges is decomposed as $P(\mX, \mW, \mY)=P(\mY \mid \mW) P(\mX \mid \mY) P(\mW)$. 

\textbf{Remark.} The decomposition suggests that a properly defined DGP for a graph consists of 1) the potential function $P(\mY\mid\mW) = \frac{1}{Z} \prod_{\xi \in \mathrm{cl}(G)} \varphi_\xi\left(\gV_\xi\right)$ where $\operatorname{cl}(\G)$ is the set of cliques of $\G$, $\gV_\xi$ is the subset of nodes related to $\xi$, and $Z$ is the partition function, 2) the emission process $P(\mX\mid\mY)$ that generate node features and 3) the structural prior $P(\mW)$ that encodes the graph structure, such as sparsity~\citep{10.1093/biostatistics/kxm045} or connectivity~\citep{pmlr-v51-kalofolias16}. 

We follow ~\citet{zhu2003semi} and consider the cliques up to rank 2, i.e., the node-wise and pair-wise potential are selected to be $\varphi_1(v) = \exp(-(d_v+\nu)\|\vy_v\|^2)$ with $\|\cdot\|$ the $L^2$ norm and $\varphi_2(u, v) = \exp(w_{uv}\vy_u^\top\vy_v)$ respectively. This leads to a multi-variate Gaussian on each column of the hidden variable $\mY_{:k}$ with zero mean and precision matrix $\Lambda = \mL + \nu \mI$ where $\mI$ is the identity matrix. The emission function is considered as another multi-variate Gaussian~\citep{pmlr-v33-liu14} which has a linear relationship, specifically $\vx_v \mid \vy_v  \sim \N (\mV \vy_v, \Sigma_\vx)$. 
We denote $\mV$ as the linear transformation matrix and $\Sigma_x$ the marginal covariance matrix for $\vx_v$, which are shared across all nodes $v\in \gV$. 

Under the assumption that the marginal noise on $\vx$ is not too strong to conceal the structure information, i.e., $\det(\Sigma_\vx) \ll \det(\Lambda^{-1})$.
we can marginalize out $\mY$ and obtain the likelihood function of $\mX$ as
\begin{equation}
\label{eq:likelihood_HMN_signals}
\begin{aligned}
    P_\mV(\mX\mid \mW) &=\frac{1}{Z}  \prod_v \varphi_1(v) \prod_{u,v}\varphi_2(u,v)
    \\
    \text{with }\varphi_1(v)&= \exp(-(\nu+d_v)\|\mV^{\dagger} \vx_v\|^2) \\
    \text{and }\varphi_2(u,v) &= \exp \left\{ w_{uv} (\vx_u^\top\mV^{\dagger \top}\mV^\dagger\vx_v)\right\},
\end{aligned}
\end{equation}
where $\mV^\dagger$ is the Moore–Penrose pseudo inverse of matrix $\mV$. We left the detailed derivation of the likelihood function in~\cref{apdx: generative_process}.

\subsection{Graph Structure Learning} 
\label{subsec_vanillaGSL}

The link between DGP under PE-MRF assumptions and GSL was introduced in ~\cite{DBLP:conf/aistats/ParkHBL17}. Here we extend the result to HMN and cast the GSL as a MAP estimation problem for the graph structure $\mW$ and the linear transformation matrix $\mV$\footnote{$\nu$ is considered as a hyperparameter.}, where,
\begin{equation}
\label{eq: map}
\begin{aligned}
&\mW^*, \mV^* =\argmax_{\mW, \mV} \log P_\mV(\mW \mid \mX) \\&=\argmax_{\mW, \mV} \log P_\mV(\mX \mid \mW) +\log P(\mW) + \log P(\mV).
\end{aligned}
\end{equation}
Current graph structure learning (GSL) algorithms~\citep{DBLP:journals/tsp/DongTFV16, pu2021learning, DBLP:conf/nips/0005YCP19} can be considered as a special case of the above MAP problem when $\mV^{\dagger}$ is rectangular orthogonal (semi-orthogonal) matrix. In such a scenario, the potential functions in~\cref{eq:likelihood_HMN_signals} reduces to $\varphi_1(v) = \exp(-(\nu+d_v)\|\vx_v\|^2)$ and $\varphi_2(u,v)=\exp \left\{ w_{uv} (\vx_u^\top\vx_v)\right\}$,
which equates to a PE-MRF.
Solving the MAP problem in~\cref{eq: map} leads to, 
\begin{equation}
\label{eq: GSL_format}
\argmin_\mW \underbrace{\sum_{uv} w_{uv} \|\vx_u -\vx_v\|^2}_{S(\mX, \mW)} - \underbrace{\mathbf{1}^\top\log \left(\mW \cdot \mathbf{1}\right) -\log P(\mW)}_{\Omega(\mW)}.
\end{equation}
The training objective consists of graph-signal fidelity term $S(\mX, \mW)$ and a structural regularizer $\Omega(\mW)$. To conclude, learning the graph topology through MAP estimation parameterized by HMN is equivalent to the GSL algorithms if 1) $\mV^\dagger$ is semi-orthogonal, 2) $\det(\Sigma_\vx) \ll \det(\Lambda^{-1})$, 3) The log-determinant of the partition function is relaxed by the lower-bound in the optimization. We left the derivation in~\cref{apdx: GSL_training_obj}.

\section{HETEROGENEOUS GRAPH STRUCTURE LEARNING}

Our work focuses on generalizing HMN to facilitate the design of DGP, and solve the HGSL problem in~\cref{eq: hetero_log_likelihood} via the link established in~\cref{sec: DGP_GSL_connection}. In this section, we will first propose a novel data-generating model in~\cref{sec:gen_model} that assists in parameterizing $P(\{\vx_v\}\mid \tW, \Theta )$, and then dive into algorithm design to solve the problem in~\cref{sec:method}. Finally we present analyses of the proposed algorithm in~\cref{sec: alg_analysis}.

\subsection{Data-generating Process for Heterogeneous Graphs}
\label{sec:gen_model}

We redefine the potential and emission functions in the DGP and obtain a novel model which we call hidden Markov networks for heterogeneous graphs (H2MN).

\paragraph{The model.}

We extend a graphical model for HG as in~\cref{fig: graphical}b. Similar to the HMN introduced in ~\cref{subsec:prob_form}, the likelihood can be decomposed into the node-wise potential $\varphi_1(v)$, the pair-wise potential $\varphi_2(v, u \mid\mB_r)$ and the signal emission process whose density function is $P(\vx_v\mid \vy_v)$. We first consider the joint PDF of node signals and labels that yields
\begin{equation}
\begin{aligned}
\label{eq:joint_likelihood}
    &P(\{\vx_v\}, \{\vy_v\}\mid \tW, \{\mB_r\}) 
    \\&=P(\{\vy_v\}\mid\tW, \{\mB_r\}) \prod_v P\left(\vx_v \mid \vy_v \right). 
\end{aligned}
\end{equation}
Similar to the HMN for homogeneous graphs, the generation density of $\{\vy_v\}$ in HGs is defined as,
\begin{equation}
\label{eq: generation_distribution}
\begin{aligned}
&P(\{\vy_v\}\mid \tW, \{\mB_r\}) \propto \prod_v \varphi_1\left(v \right) \prod_{u,v,r} \varphi_r \left(u, v\right)
\end{aligned}
\end{equation}
where the node-wise potential function is unchanged as $\varphi_1(v) = \exp(-(d_v+\nu)\|\vy_v\|^2)$, and the edge-wise potential function is replaced by a bi-linear product dependent on the relation type $\varphi_r(u, v) = \exp(w_{uvr}\vy_u^\top\mB_r\vy_v)$. 

To link label variable $\vy_v$ with the signal variable $\vx_v$, we assume that $P(\vx_v \mid \vy_v)$ has a mean that is a linear function of $\vy_v$, and a covariance $\sigma^2\mI$ that is independent of $\vy_v$, which gives,
\begin{equation}
\label{eq: emission_func}
\vx_v \mid \vy_v \sim \N\left(\vx_v \mid \mV_{\phi(v)} \vy_v, \sigma^2\mI \right).
\end{equation}
The linear transformation matrix $\mV_{\phi(v)}\in \R^{K\times |C_{\phi(u)}|}$ is determined by the node type $\phi(v)$. We then substitute~\cref{eq: generation_distribution} and~\cref{eq: emission_func} into~\cref{eq:joint_likelihood}, marginalize out $\{\vy_v\}$ and obtain the likelihood for the node signals, 
\begin{equation}
\label{eq:likelihood_signals}
\begin{aligned}
    &P(\{\vx_v\}\mid \tW, \{\mB_r\}) \propto \prod_v \varphi_1(v) \prod_{u,v,r} \varphi_r(u,v) \\
    & \text{with } \varphi_1(v) = \exp(-(\nu+d_v)\|\mU_{\phi_v} \vx_v\|^2) \\
    &\text{and }\varphi_r(u,v) = \exp \left\{ w_{uvr} (\vx_u^\top\mU_{\phi(u)}^\top \mB_r \mU_{\phi(v)}\vx_v)\right\},
\end{aligned}
\end{equation}
where $\mU_{\phi(u)} \in \R^{|C_{\phi(u)}|\times K}$ is the Moore–Penrose pseudo inverse of matrix $\mV_{\phi(u)}$. The detailed derivations can be found in~\cref{subsec: DGM_HG}

\textbf{Remark.} Though we use a pair-wise HMN to approximate the overall joint probability distribution, this approximation preserves the interdependencies among edges with various relation types across the graph, even when they are not directly connected via a single node. The proof can be found in~\cref{apdx: expressiveness}.

\subsection{Algorithm Design}
\label{sec:method}

\textbf{The optimization objective.} To further simplify the estimation, we conduct a low-rank approximation by\footnote{$\mU_{\phi(u)}^\top \mB_r \mU_{\phi(v)}$ is positive semi-definite if we consider the nodes for different types share the same signal space.} $\mM_r \mM_r^\top \approx \mU_{\phi(u)}^\top \mB_r \mU_{\phi(v)}$, where we specify a relation-wise matrix $\mM_r\in \gR^{K \times K^\prime}$ that depends only on $r$ since $\phi{(u)}$ and $\phi{(v)}$ can be directly determined when $r \in \mathcal{R}_{\phi(v),\phi(u)}$ is given. Our empirical study shows that a rank-1 relation embedding $\ve_r \in \gR^K$ is sufficient in the approximation. Thus, the training objective can be re-formalized as (derivation in~\cref{apdx: training_obj_hgsl}),
\begin{equation}
\begin{aligned}
\label{eq:training_obj}
\argmin_{\tW, \mE} &\sum_{v\leq u, r} w_{vur}\| \ve_r^\top\cdot(\vx_v- \vx_u)\|^2 + \Omega(\tW) + \Omega (\mE)
\end{aligned}
\end{equation}
where the relation-wise vectors $\ve_r$ are stacked as $\mE = [\ve_1, \ldots, \ve_{|\mathcal{R}|}]$. One can see that the first term, $S(\mX, \mE, \tW)=\sum_{u, v, r} w_{vur}\| \ve_r ^\top\cdot (\vx_v- \vx_u)\|^2$, promotes signal smoothness similar to that in the graph structure learning literature~\citep{pu2021learning, DBLP:journals/tsp/DongTFV16} as in~\cref{eq: GSL_format}. 

Since the problem requires learning both the \textit{graph structure} $\tW$ and \textit{relation embeddings} $\mE$ that is not jointly convex, we adopt an alternating optimization scheme to solve it: $\tW$ is optimized with $\mE$ fixed, and $\mE$ is optimized with $\tW$ fixed. For notation consistency, we will use $\E^\prime$ to denote the possible connections in a heterogeneous graph. Note that this is not the combination of all possible nodes and relation types that shapes a $\gV \times \gV \times \gR$ space as node pairs with certain types can only be connected by specific edge types.

\begin{algorithm}[t!]
 \caption{Heterogeneous Graph Structure Learning with Alternating Optimization.}\label{alg:RE}
\begin{algorithmic}
\renewcommand{\algorithmicrequire}{\textbf{Input:}}
     \renewcommand{\algorithmicensure}{\textbf{Output:}}
\REQUIRE{Node signals $\{\vx_v\}_{v=1}^{|\mathcal{V}|}$ and types $\{\phi(v)\}_{v=1}^{|\mathcal{V}|}$; \\ Relation type set $\mathcal{R}$; Maximum steps $T$.}
\ENSURE{The graph weight $\vw$}; Embeddings $\ve_r, \forall r \in \mathcal{R}$ \;
 \textbf{Init}: Initialize $\ve_r^0 =  {\mathbf{1^T}}/{K}, \forall r\in \mathcal{R}$; Initialize $\vw^0$ randomly; Initialize $t=0$\; 
 \While{t $<$ $T$}{
 \tcc{Graph structure learning step}
 Calculate the smoothness vector $\vz$\;
 Optimize $\vw^{t+1}$ based on~\cref{eq: optimization_obj}\;
 \tcc{Relation embedding update Step}
 \IF{Update Method == Gradient Descent}
 \STATE $\ve_r^{t+1} = \text{GD}(\ve_r^t)$ according to~\cref{eq: relation_emb_update}\;
 \ELSIF{Update Method == Iterative Reweighting}
 \STATE Update $\ve_r^{t+1}$ based on~\cref{eq: FR}\;
 \ENDIF \;
 t = t+1\;
 }
\end{algorithmic} 
\label{alg: main}
\end{algorithm}

\paragraph{Graph structure learning step.} The first sub-optimization problem is to find $\tW$ that minimizes the training objective in~\cref{eq:training_obj} with a fixed $\mE$,
\begin{equation}
\begin{aligned}
&\argmin_{\tW} S(\mX, \mE, \tW) + \Omega(\tW).
\end{aligned}
\end{equation}
Since we focus on undirected graphs (i.e., $\tW$ symmetric), we only need to learn the upper triangle part of the tensor, i.e., the vectorized weights $\vw \in \mathbb{R}_+^{|\mathcal{V}|\cdot (|\mathcal{V}|-1)\cdot |\mathcal{R}|/2}$. 
We reparameterize the training task as in \cite{pu2021learning} and obtain $\sum_{\{v, u, r\}\in \E^\prime} w_{vur}\|\ve_r\circ(\vx_v- \vx_u)\|^2 = \|\vw \odot \vz\|$, where 
$\vz \in \mathbb{R}_+^{|\mathcal{V}|\cdot (|\mathcal{V}|-1)\cdot |\mathcal{R}|/2}$ is the half-vectorization of the tensor $\| (\mX\otimes\mathbf{1}- \mathbf{1}\otimes\mX) \otimes \mE\|_F^2$ with $\otimes$ the outer product. 
In addition, inspired by classical GSL methods~\citep{DBLP:conf/aistats/Kalofolias16, DBLP:journals/tsp/DongTFV16}, we choose $\Omega(\tW)$ to consist of a log barrier regularizer on node degrees and a $L^1$ norm regularizer on the weights to promote the connectivity and the sparsity of the graph respectively. 
The objective is reformulated as, \begin{equation}
\label{eq: optimization_obj}
\argmin_\vw \|\vw \odot \vz\|^2 - \alpha \mathbf{1}^{\top} \log (\mathcal{T} \vw) +\beta\|\vw\|_1 + \mathcal{I}_{\vw>0},
\end{equation}
where $\odot$ is the element-wise product and $\mathcal{T}$ is a linear operator that transforms $\vw$ into the vector of node degrees such that $\mathcal{T}\vw = (\sum_{r = 1}^{|\mathcal{R}|} \tW_{::r})\cdot \mathbf{1}$. We solve \cref{eq: optimization_obj} by alternating direction method of multipliers (ADMM)~\citep{pu2021learning} or primal-dual splitting algorithms (PDS)~\citep{DBLP:conf/aistats/Kalofolias16}. 

\paragraph{Relation embedding update step.}
\label{sec: opt_alg}

The second sub-problem handles the optimization of relation embeddings $\ve_r$. Considering $\Omega(\mE)$ as an elastic norm~\cite{10.1111/j.1467-9868.2005.00503.x}, ~\cref{eq:training_obj} becomes
\begin{equation}
\label{eq: relation_emb_update}
\begin{aligned}
\argmin_{\{\ve_r\}} \quad &S(\mX, \mE, \tW) + \lambda_1 \|\mE\|^2 + \lambda_2 \|\mE\|_1.
\end{aligned}
\end{equation}
$\lambda_1$ and $\lambda_2$ are hyper-parameters. Intuitively, $\ve_r$ can be directly optimized by gradient descent, and we denote this approach as HGSL-GD. However, in our experiments, we found this unstable and easily stuck into a sub-optimum. Thus, a more efficient and stable solution is developed through analytically solving~\cref{eq: relation_emb_update} w.r.t. $\ve_r$. The detailed derivation is left in~\cref{apex: der_relation_emb}. We denote this approach iterative reweighting (HGSL-IR): intuitively this solution assigns higher weights for relation $r$ to dimensions that express higher dimension-wise similarity on the graph learned at iteration $t$ (i.e., $w_{vur}^t$): 
\begin{equation}
\begin{aligned}
\ve_{r, k}^{t+1} &= \frac{\lambda_2}{2\lambda_1}(\frac{1}{\lambda_1}\sum_{\{v, u, r\} \in \E^\prime_r} w_{vur}^t~ \vx_{v,k} \cdot \vx_{u,k}-2)\\
&= \lambda_1^\prime \sum_{\{v, u, r\} \in \E^\prime_r} w_{vur}^t~ \vx_{v,k} \cdot \vx_{u,k}-\lambda_2^\prime
\label{eq: FR}
\end{aligned}
\end{equation}
where $\E^\prime_r$ is the set of possible edges of type $r$ and the subscript $k$ denotes the $k$-th dimension, $\lambda_1^\prime = \frac{\lambda_2}{2\lambda_1^2}$ and $\lambda_2^\prime = \frac{\lambda_2}{\lambda_1}$. It is noted that $\lambda_1$ cannot be chosen to be large to avoid a trivial solution (e.g. $\ve_r=0$). The overall process is described in \cref{alg: main}. 

\subsection{Algorithm Analysis}
\label{sec: alg_analysis}

The hierarchical nature of the DGP allows us to evaluate algorithm performance from two perspectives: the potential functions linking node label generation to the underlying graph structure, and the emission functions governing signal generation. We will demonstrate how the connectivity matrix within the potential function relates to homophily and how the signal generation process in the emission function affects edge type distinguishability, both of which influence model performance. Based on this understanding, we introduce \hypertarget{impact_factor}{two factors} that impact algorithm performance.

\paragraph{Homophily on heterogeneous graphs:} Homophily graphs~\citep{doi:10.1146/annurev.soc.27.1.415} suggest that connections between similar entities occur at a higher probability than among dissimilar entities. This translates to $P(\mW_{uv} =1 \mid \vy_u= \vy_v) > P(\mW_{uv} =1 \mid \vy_u \neq \vy_v)$, 
where $\vy_u$ and $\vy_v$ are the labels for nodes $u$ and $v$, and the homophily ratio is linked to the ratio of the two values~\citep{DBLP:conf/iclr/0001LST22} with the relationship suggested in~\cref{apdx: homophily_proof}. We will generalize the concept of homophily in HGs through the connection probability matrix $\{\mB_r\}$. 
To do so, a ``representative'' connection is defined between two classes $p^*$ and $q^*$ of nodes among relation type $r$ if it gives the highest probability of connection $\mB_{r}[p^*, q^*]$, i.e., 
\begin{equation}
\label{eq: homophily_heteorograph}
\mB_{r}[p^*, q^*] -  \sum_{p \neq p^* \cap q \neq q^*} \mB_{r}[p, q] > 0, \forall r.
\end{equation}
The inequality yields the definition of homophily ratio,
\begin{equation}
\label{eq: hg_homo_ratio}
\text{HR}(\G, r) = \frac{\mB_{r}[p^*, q^*] }{\sum_{p \neq p^* \wedge q \neq q^*} \mB_{r}[p, q]}.
\end{equation}
In~\cref{apdx: homophily_proof} we prove that inequality in~\cref{eq: homophily_heteorograph} is a sufficient condition for guaranteeing~\cref{alg:RE} a meaningful solution\footnote{Avoiding non-convergence solution such as infinitely maximizing $\tW$, or trivial solutions such that $\tW = 0$ }. 
However, in practice, the value in~\cref{eq: hg_homo_ratio} is difficult to compute, because labels are typically recorded for only one node type, rather than all node types while constructing graph datasets for node classification tasks. Therefore, we consider a new metric, named relaxed homophily ratio, in the following paragraph.

\textbf{[Relaxed Homophily Ratio (RHR)]}~\citep{DBLP:conf/www/GuoDBFMCH0Z23} is defined based on how the labels are similar along the meta-paths~\citep{DBLP:journals/corr/abs-2011-14867}. A meta-path $\Phi$ is a path template following a specific sequence of node and relation types like $A_1 \stackrel{R_1}{\longrightarrow} A_2 \stackrel{R_2}{\longrightarrow} \cdots \stackrel{R_{L-1}}{\longrightarrow} A4_L$ with node types $A_1, \cdots, A_{L} \in \mathcal{A}$ and edge types $R_1,\cdots, R_{L-1} \in \mathcal{R}$. When $A_1 = A_L$, it is possible to match the node and edge types along each path $\gP$ in $\gG$ with meta-path $\Phi$ and construct a new graph $\gG_\Phi$ by connecting the source and end nodes for all $\gP$. The RHR is defined on $\gG_\Phi$ by measuring the label ($y_u$) similarity within the same node type: 
\begin{equation}
\label{eq: relaxed_hr}
\text{RHR}\left(\mathcal{G}_{\Phi}\right)=\frac{\sum_{\{u, v\} \in \mathcal{E}_{\Phi}} \mathbb{I}\left(y_u=y_v\right)}{\left|\mathcal{E}_{\Phi}\right|},
\end{equation}
We show in~\cref{sec: rbst_analysis} that the RHR is positively related to the model performance. 

\paragraph{Edge type distinguishability:}
We now examine the property of the signal generation function and its impact on the solution. Intuitively illustrated in~\cref{fig: gen_smooth}, the data-fidelity term in~\cref{eq:training_obj} can be considered as a \emph{reweighted smoothness} scheme that first measures dimension-wise smoothness, and then integrates it by emphasizing specific signal dimensions according to $r$. Using the movie review dataset again as an example, while determining whether a `star in'-typed edge should be formed between two actor nodes, the model should put larger weights on the signal dimensions that represent the `genres', but weigh less irrelevant ones such as `company affiliation' or `date'. 

Thus, an important property that supports the learning is that the nodes connected with different relation types (inherently different node labels) would exhibit similarity in different dimensions of features. This can be measured by the following quantity, namely the smoothest-dimension overlapping ratio (SDOR). 

\textbf{[Smoothest-Dimension Overlapping Ratio (SDOR)]} We first define the dimension-wise smoothness as $S(\mX_{:k}, \tW_{::r}) = \sum_{\{v,u,r\} \in \E} w_{vur}\|\x_{v,k}-\x_{u,k}\|_2^2$ for relation $r$ and dimension $k$. The top-$M$ smooth dimensions for $r$, $\K^M(r)$ are calculated by ranking dimension-wise smoothness across $k\in[K]$. For each relation pair $(r, r^\prime)$, the SDOR is obtained by counting the overlapping dimensions in $\K^M(r)$ and $\K^M(r^\prime)$:
\begin{equation}
\text{SDOR}(r, r^\prime) = \frac{|\K^M(r)\cap \K^M(r^\prime)|}{|\K^M(r)\cup \K^M(r^\prime)|}.
\end{equation}
The SDOR reflects how different two relation types are in terms of exhibiting smoothness in signal generation. If relation pairs $(r, r^\prime)$ exhibit a high SDOR, our algorithm will converge into a solution that has similar embeddings for $r$ and $r^\prime$, leading to difficulties in distinguishing between the two relation types. 
In~\cref{sec: rbst_analysis}, we will empirically demonstrate the correlation between SDOR and algorithm performance.

\section{RELATED WORKS}

\textbf{Multiple Graphical Models Estimation.} A line of research has extended precision matrix estimation methods such as the graphical Lasso to heterogeneous data, referred to as multiple graphical model estimation (MGME)~\citep{DBLP:conf/nips/GanYNL19,DBLP:journals/jmlr/MaM16,DBLP:journals/jmlr/HaoSLC17}. Examples include group graphical lasso~\citep{DBLP:conf/icml/JacobOV09} and fused graphical Lasso~\citep{danaher2014joint}. The concept of heterogeneous data in MGME assumes that heterogeneity exists across multiple separate graphical models (termed as groups), while our work considers a heterogeneous graph in a different sense. In contrast to MGME methods, we assume there is a single underlying graph that contains multiple types of nodes and relationships, which motivates us to design HGSL algorithms to ensure the general property, such as connectivity and sparsity, of the entire graph.

\textbf{Supervised HGSL.} Another branch of research focuses on learning latent graph structures from supervision signals, such as existing links or downstream task objectives~\citep{DBLP:journals/mia/ZaripovaCKABN23, DBLP:conf/iclr/BattiloroSTBSL24}. Here, the goal is to construct a graph that benefits a specific downstream task. A handful of studies have also explored applying this supervised graph structure learning approach to heterogeneous graphs~\citep{DBLP:conf/aaai/ZhaoWSHSY21,DBLP:conf/kdd/ZhaoWYY23}. In contrast, our work addresses unsupervised GSL, where the aim is to model the interaction between the graph and the node features so that, if the graph is unavailable, it can be reconstructed from the node features in a way that preserves its key properties.

\section{EXPERIMENTS}

We evaluate the HGSL algorithm through quantitative and qualitative experiments. We first outline the experimental setup in \cref{sec: exp_setup}, followed by a quantitative assessment of model performance in \cref{sec: quan_results}. Next, we analyze the impact of previously mentioned factors on HGSL performance in \cref{sec: rbst_analysis} and demonstrate how our model reveals interesting relationships in real-world problems in \cref{sec: quali_results}.

\subsection{Experimental Setup}
\label{sec: exp_setup}

\subsubsection{Datasets}
\paragraph{Synthetic datasets.} The synthetic dataset construction consists of 3 phases: 1) \textbf{Graph backbone generation.} We followed the process in \cite{pu2021learning} and used the stochastic block model and Watts–Strogatz model to generate the backbone with 20-100 nodes that encode the graph structure. 2) \textbf{Node type generation.} We traverse the graph by breadth-first search and generate the node types following the rule defined in a meta template (network schema~\citep{Shi2022}) that encodes how nodes/edges with different types are connected.
\textbf{3) Feature generation.} We then generate node signals that satisfy the probability distribution as follows.
\begin{equation}
\label{eq: generation_process}
p(\mX, \mE \mid \tW  ) \propto \exp{}(-\frac{1}{\sigma}\sum_{\{v, u, r\} \in \E^\prime} w_{vur}\| \ve_r\cdot(\vx_v- \vx_u)\|_2^2).
\end{equation}

\textbf{Real-world datasets}: We consider IMDB~\citep{DBLP:conf/www/GuoDBFMCH0Z23} and ACM~\citep{DBLP:conf/kdd/LvDLCFHZJDT21} datasets for quantitative evaluation. IMDB is a movie review dataset with node types including directors (D), actors (A), and movies (M) and with signals as 3066-D bag-of-words representation. ACM is an academic dataset that contains papers (P), authors (A), and subjects (S). Signals correspond to the 1902-D bag-of-words representation of the keywords in diverse research areas. Detailed statistics are presented in \cref{tab: lpresult}. To augment more graph instances for a comprehensive evaluation, we subsample the dataset to generate smaller graphs with number of nodes ranging from 50-200 following the strategy in \cite{DBLP:conf/www/HuDWS20}. 

We also perform a qualitative experiment on YahooFinance~\citep{yahoo_finance} financial dataset to uncover the relationships among S\&P 100 companies with node types representing company sectors (health, finance, technology). Using daily stock returns as signals, the aim is to reveal connectivity patterns, represented by edge types, across different company sectors.

\subsubsection{Metrics}

The model performance is evaluated quantitatively on their ability to identify edge type and to recover edge weights. The edge type identification error is measured by the area under the curve (AUC) where we view ``no edge'' as a class and consider a classification problem on the edge types. The edge weight recovery error is measured by the mean squared error for graph recovery (GMSE) introduced by ~\cite{pu2021learning},
\begin{equation}
\mathrm{GMSE}=\frac{1}{|\gV|\times |\gV| \times |\mathcal{R}|} \sum_{\{u,v,r\}} \frac{\left\|\hat{w}_{uvr}-w_{uvr}\right\|^2}{\left\|w_{uvr}\right\|^2}
\end{equation}

\subsection{Quantitative Results}
\label{sec: quan_results}

\begin{table*}
\centering
\resizebox{\textwidth}{!}{
\begin{threeparttable}
\renewcommand\arraystretch{0.6}
\center
\begin{tabular}{lccccccccc}
\toprule 
\multirow{2}{*}{\bf Model}
& \multicolumn{2}{c}{\bf Synthetic Dataset} & \multicolumn{2}{c}{\bf IMDB (HR: 0.51) } & \multicolumn{2}{c}{\bf ACM (HR: 0.64)} \\
\cmidrule(r){2-3} \cmidrule(r){4-5} \cmidrule(r){6-7} 
& AUC & GMSE & AUC & GMSE & AUC & GMSE  \\

\midrule

GGL~\citep{DBLP:conf/icml/JacobOV09}
&0.63 $\pm$ 0.03& 0.06 $\pm$ 0.02 
&0.60 $\pm$ 0.06 &0.5 $\pm$ 0.02
&0.68 $\pm$ 0.05& \textbf{0.02 $\pm$ 0.01} \\

FGL~\citep{danaher2014joint}
&0.62 $\pm$ 0.03 &0.05 $\pm$ 0.01	&0.73 $\pm$ 0.06	&
\textbf{0.03 $\pm$ 0.03 }& 0.59 $\pm$ 0.03	&0.03 $\pm$ 0.00  \\

\midrule

GSL~\citep{DBLP:journals/tsp/DongTFV16} &0.61 $\pm$ 0.04 &0.04 $\pm$ 0.00 & 0.75 $\pm$ 0.06 &0.30 $\pm$ 0.04	&0.56 $\pm$ 0.06 & 0.09 $\pm$ 0.08\\
GSL~\citep{DBLP:conf/aistats/Kalofolias16}& 0.58 $\pm$ 0.02	& 0.04 $\pm$ 0.02 & 0.74 $\pm$ 0.08 & 0.27 $\pm$ 0.10 & 0.57 $\pm$ 0.04 & 0.14 $\pm$ 0.03\\
GSL~\citep{pu2021learning}& 0.65 $\pm$ 0.01& 0.03 $\pm$ 0.00& 0.74 $\pm$ 0.04& 0.29 $\pm$ 0.08& 0.65 $\pm$ 0.06& 0.07 $\pm$ 0.10\\
\midrule
HGSL-IR (Ours)& \textbf{0.83 $\pm$ 0.04}& \textbf{0.02 $\pm$ 0.01}&\textbf{0.81 $\pm$ 0.07}& 0.07 $\pm$ 0.05&\textbf{0.73 $\pm$ 0.02}& 0.12 $\pm$ 0.07 \\
\bottomrule
\end{tabular}
\caption{Quantitative experimental results on heterogeneous graph structure learning compared to covariance matrix recovery algorithms and graph structure learning algorithms for homogeneous graphs.}
\label{tab: additional_results}
\begin{tablenotes}
\item $^*$ Experimental results are evaluated over 30 trials and the mean/standard deviation is calculated.
\end{tablenotes}
\end{threeparttable}
}
\end{table*}

Due to the lack of appropriate baselines as no prior algorithm can identify the edge types, we compare against MGME methods, specifically group graphical lasso (GGL)~\citep{DBLP:conf/icml/JacobOV09} and fused graphical Lasso (FGL)~\citep{danaher2014joint}, and conventional graph structure learning (GSL) algorithms\footnote{Following the implementation in~\cite{pu2021learning}.} ~\citep{DBLP:conf/aistats/Kalofolias16,pu2021learning} in terms of edge weights recovery and binary edge identification (omitting the types). The results are reported in \cref{tab: additional_results}.
With our HGSL algorithm with iterative reweighting (HGSL-IR), a consistent improvement is found in all the datasets in terms of both AUC and GMSE. This suggests the efficacy of our algorithm in both tasks. The improvement of AUC in synthetic datasets (avg +18.48$\%$) is relatively larger than the real-world datasets (avg +10.88$\%$), which suggests the more challenging nature of the real-world experiments.

We also test the HGSL with gradient descent (GD) to demonstrate the efficacy of our algorithm design in~\cref{sec:method}, where the results are reported in~\cref{tab: new_algorithm_result}. Though a better AUC is found in most datasets, the improvement is rather marginal and the GMSE is much larger. This shows our HGSL-IR can lead to better performance.

\subsection{Robustness Analysis}
\label{sec: rbst_analysis}

\begin{figure}[t!]
    \centering
    \begin{subfigure}[t]{0.44\textwidth}
        \centering
        \includegraphics[width=1\textwidth]{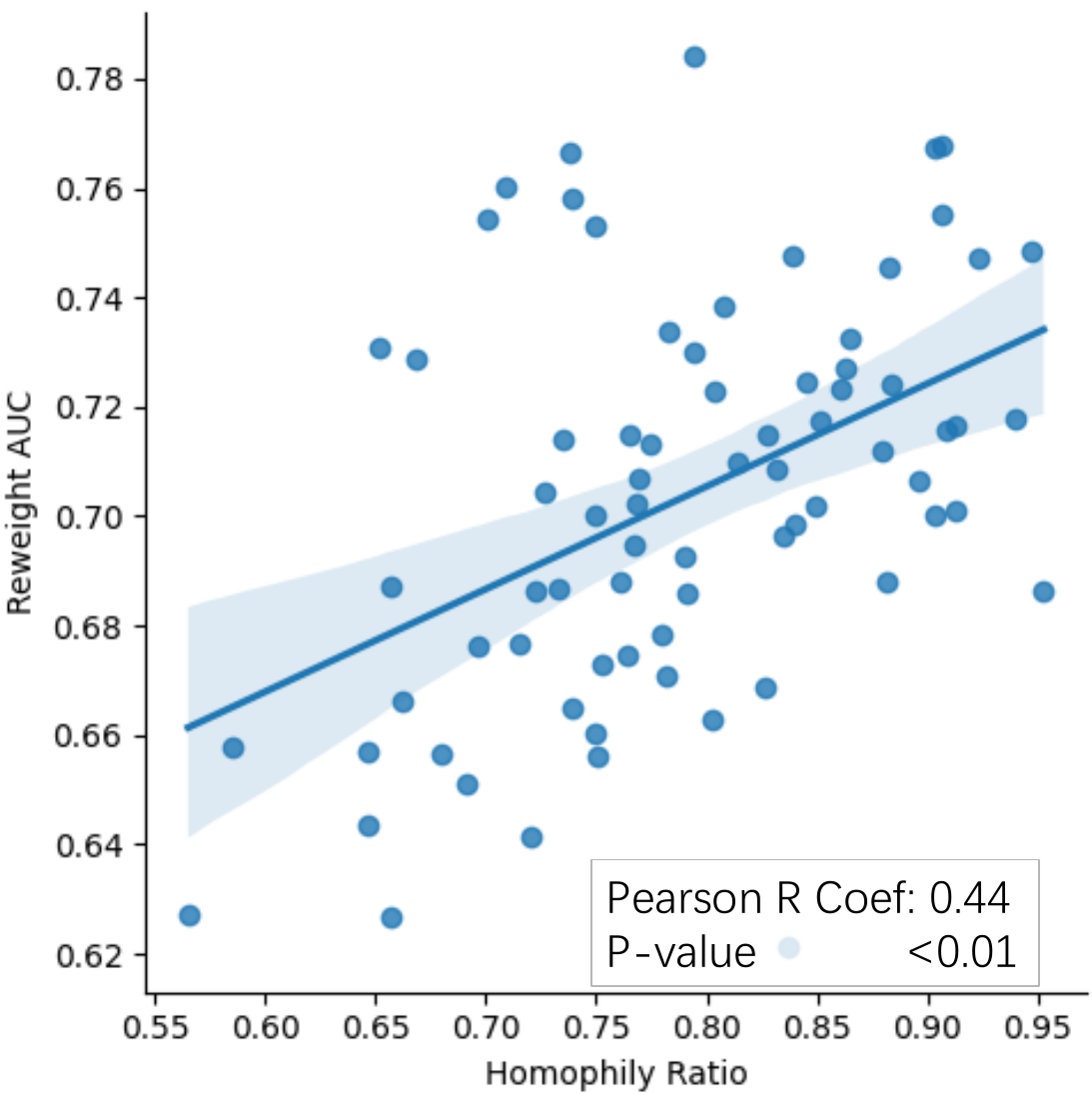}
        \caption{ACM (P-Coef: 0.44)}
    \end{subfigure}%
    ~
    \begin{subfigure}[t]{0.44\textwidth}
        \centering
        \includegraphics[width=1\textwidth]{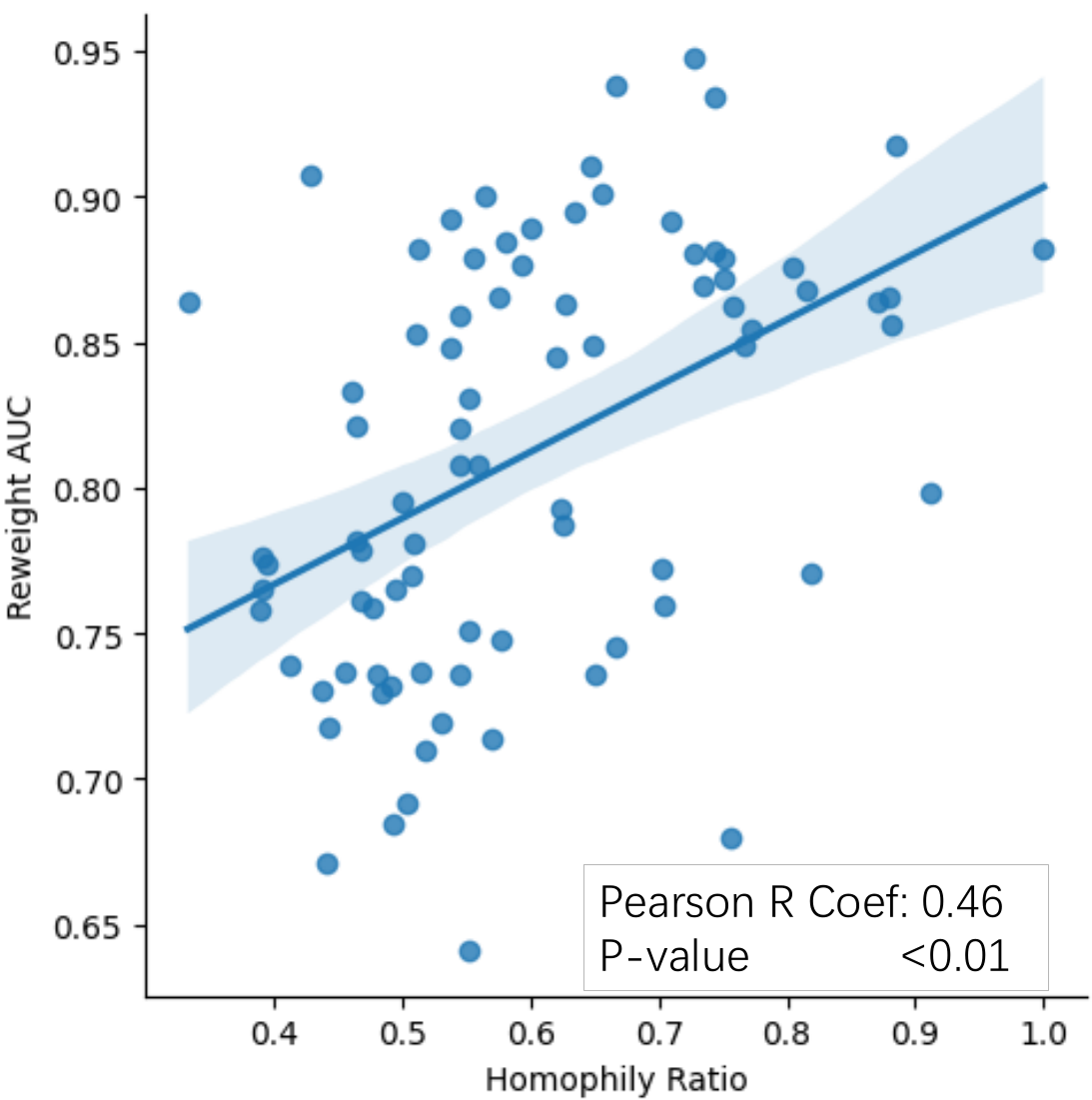}
        \subcaption{IMDB (P-Coef: 0.46)}
    \end{subfigure}
    \caption{The Pearson correlation test between relaxed homophily ratio and AUC.}
    \label{HR_result}
\end{figure}

\paragraph{Relaxed homophily ratio.} 
We are interested in how the algorithm performance is related to the homophily ratio. From our theoretical analysis in~\cref{sec: alg_analysis}, we hypothesize that a lower RHR hinders performance. Thus, we record the RHR of all the subgraphs and report the Pearson correlation coefficients between RHR and corresponding AUC as shown in \cref{HR_result}. According to the results, we can conclude that the homophily ratio is positively correlated with AUC, which validates our analysis.

\begin{figure}[t!]
  \includegraphics[scale=0.130]{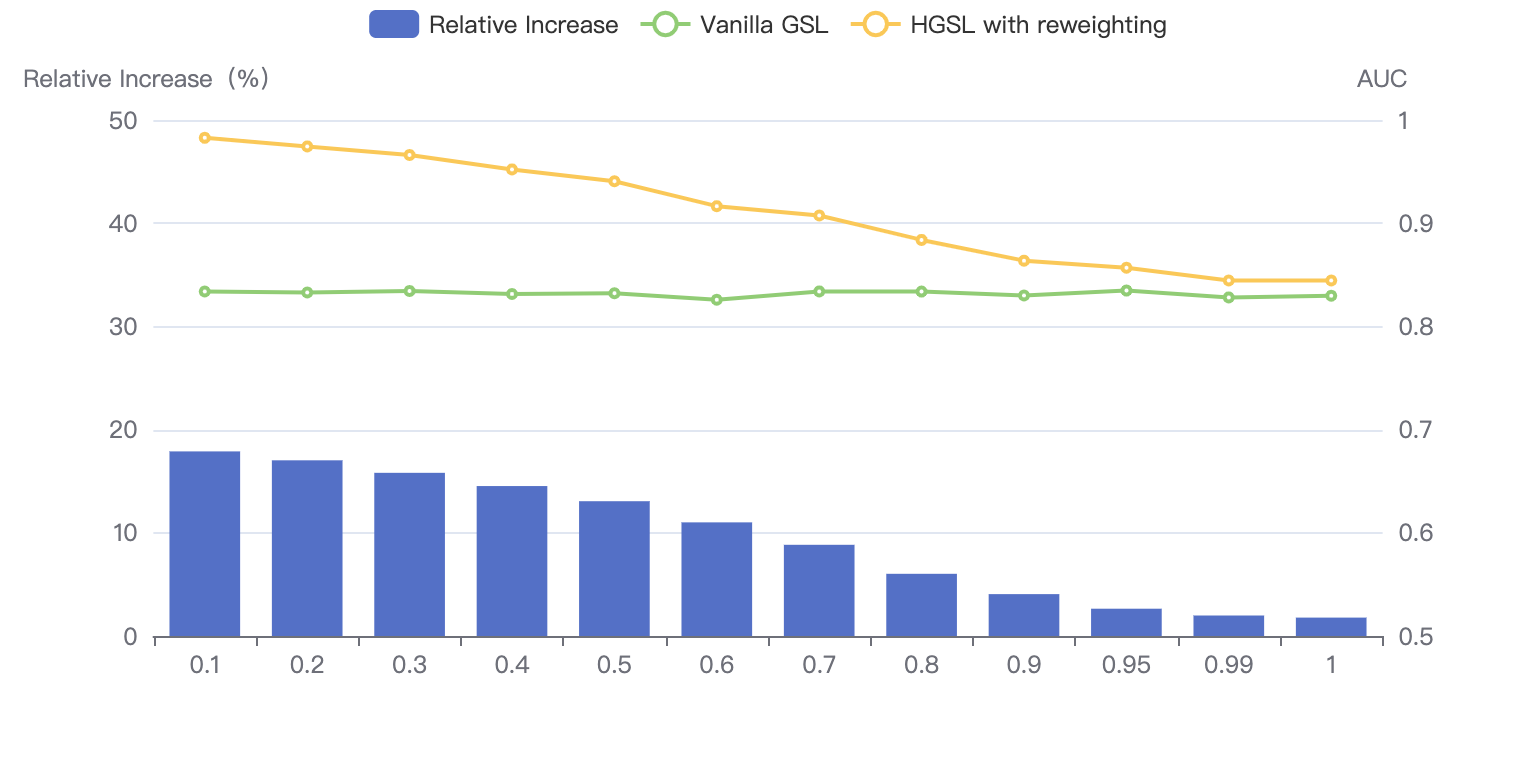}
  \caption{The SDOR and model performance.}
  \label{fig: ovp_rate}
\end{figure}

\paragraph{Smoothest-dimension overlapping ratio.}
To understand the behavior of our algorithm against high \textbf{SDOR}, we fix the number of relation types $|\mathcal{R}|=2$ in our synthetic experiment, and manually adjust the SDOR from 0 to 1. The higher SDOR is thought to hinder the distinguishability of relation types. 
We illustrate the AUC of the vanilla GSL and HGSL algorithms in \cref{fig: ovp_rate}, and calculate its relative increase. It is clear that the performance of HGSL is significantly affected when the SDOR increases, which suggests that a higher SDOR would hinder the distinguishability of edge types. However, the HGSL algorithm is robust until the SDOR is increased to approximately 0.7, which suggests the applicability of our algorithm in most real-world datasets, as compared with the SDOR statistics in~\cref{tab: lpresult}.

\subsection{Qualitative Results}
\label{sec: quali_results}
\begin{figure}
        \centering
        \begin{subfigure}[b]{0.49\textwidth}
            \centering
            \includegraphics[width=\textwidth]{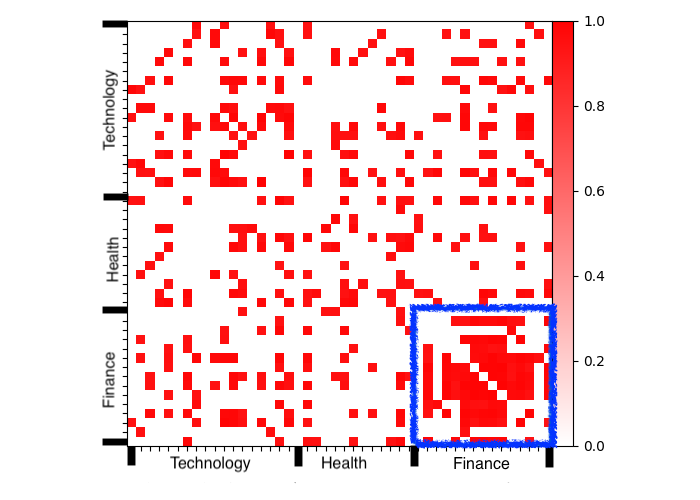}
            \caption[Network2]%
            {{\small Connections with type finance-to-finance.}}    
            \label{fig:mean and std of net14}
        \end{subfigure}
        \hfill
        \begin{subfigure}[b]{0.49\textwidth}  
            \centering 
            \includegraphics[width=\textwidth]{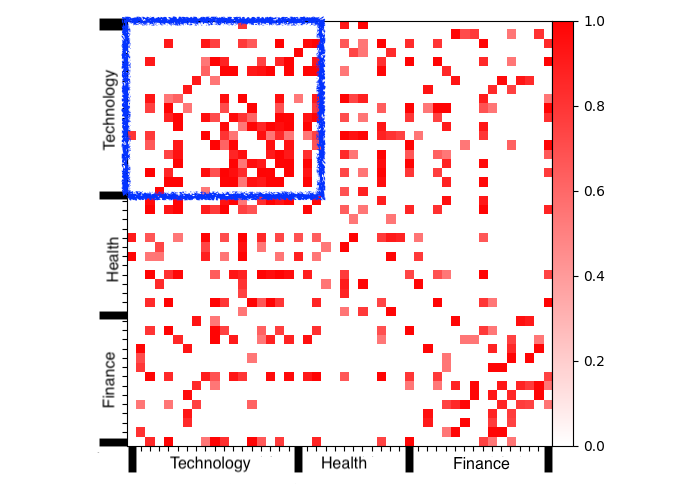}
            \caption[]%
            {{\small Connections with type technology-to-technology.}}    
            \label{fig:mean and std of net24}
        \end{subfigure}
        \vskip\baselineskip
        \begin{subfigure}[b]{0.49\textwidth}   
            \centering 
            \includegraphics[width=\textwidth]{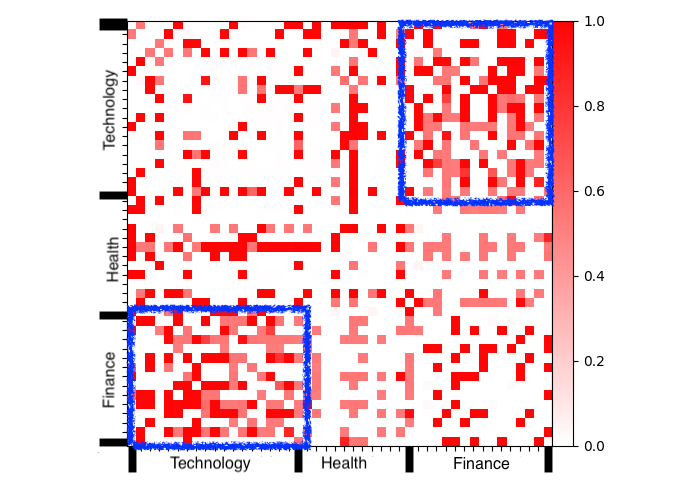}
            \caption[]%
            {{\small Connections with type technology-to-finance.}}    
            \label{fig:mean and std of net34}
        \end{subfigure}
        \hfill
        \begin{subfigure}[b]{0.49\textwidth}   
            \centering 
            \includegraphics[width=\textwidth]{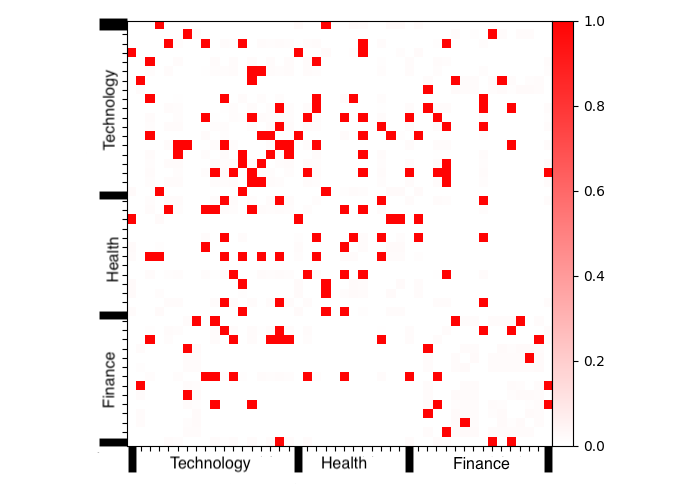}
            \caption[]%
            {{\small Connections with type health-to-technology.}}    
            \label{fig:mean and std of net44}
        \end{subfigure}
        \caption{\small The different types of connections recovered by the HGSL algorithm.} 
        \label{fig:sp100_results}
\end{figure}

We apply the HGSL algorithm to a financial dataset to recover different types of connections among S\&P 100 companies. We use the daily returns of stocks obtained from YahooFinance. Figure ~\ref{fig:sp100_results} visualized the estimated relation-wise graph adjacency matrix where sectors sort the rows and columns. The heatmap clearly shows that two stocks in the same sectors are likely to behave similarly. And most interestingly, the stock prices of companies coming from the technology and finance sectors exhibit a strong relationship. This suggests the ability of our HGSL algorithm to reveal different relationships in real-world network systems.

\section{DISCUSSION AND FUTURE WORK}

In this study, we propose an HGSL framework based on a data-generating process (DGP) for heterogeneous graphs. The method is grounded on assumptions such as signals for different node types existing in the same space and cliques being up to rank-2. However, it can easily be generalized to accommodate more complex cases with varying signal dimensions and higher-order cliques. Nonetheless, model-based design for DGP may struggle with real-world heterogeneous graphs that possess intricate structures deviating from the design. In such cases, leveraging deep learning-based generative models can be beneficial. Our future work will aim to enhance the DGP by utilizing neural networks to parameterize the potential and emission functions and developing optimization algorithms to learn the DGP. Furthermore, utilizing the DGP in scenarios with downstream tasks, such as node classification and link prediction, is a promising direction to explore. Examples along this line include understanding the performance of heterogeneous graph ML models through the compatibility between the DGP and the model architecture~\citep{DBLP:conf/iclr/KaurK023}, and informing the design of deep graph networks through understanding the DGP for graph-structured data~\citep{DBLP:conf/nips/WeiYJB022}.

\subsubsection*{Acknowledgements}

K.J. was supported by the UKRI Engineering and Physical Sciences Research Council (EPSRC) [grant number EP/R513143/1]. X.D. acknowledges support from the Oxford-Man Institute of Quantitative Finance and the EPSRC (EP/T023333/1). The authors would like to thank the anynomous reviewers for their helpful comments and insights. 

\bibliography{main}
\bibliographystyle{apalike}


\section*{Checklist}

 \begin{enumerate}

 \item For all models and algorithms presented, check if you include:
 \begin{enumerate}
   \item A clear description of the mathematical setting, assumptions, algorithm, and/or model. [Yes. The mathematical settings and assumptions are described in Sec. 2. The algorithm and model design are in Sec.4.]
   \item An analysis of the properties and complexity (time, space, sample size) of any algorithm. [Yes. We include the algorithm analysis in Appendix E.]
   \item (Optional) Anonymized source code, with specification of all dependencies, including external libraries. [No. This is a rather theoretical paper and the code for toy experiments can be provided upon request.]
 \end{enumerate}

 \item For any theoretical claim, check if you include:
 \begin{enumerate}
   \item Statements of the full set of assumptions of all theoretical results. [Yes. The assumptions for theoretical results are provided together with the statements.]
   \item Complete proofs of all theoretical results. [Yes. For every claim we made in the theoretical part, we provide the corresponding derivations in appendix.]
   \item Clear explanations of any assumptions. [Yes. We provide more explanations for the theories in the appendix.]     
 \end{enumerate}

 \item For all figures and tables that present empirical results, check if you include:
 \begin{enumerate}
   \item The code, data, and instructions needed to reproduce the main experimental results (either in the supplemental material or as a URL). [No. The code for toy experiments can be provided upon request.]
   \item All the training details (e.g., data splits, hyperparameters, how they were chosen). [Yes. We provide such experiment settings in Sec. 6.]
         \item A clear definition of the specific measure or statistics and error bars (e.g., with respect to the random seed after running experiments multiple times). [Yes. We reported the standard deviations together with the mean to illustrate statistical significance.]
         \item A description of the computing infrastructure used. (e.g., type of GPUs, internal cluster, or cloud provider). [No. The experiments are reproducible through any CPUs.]
 \end{enumerate}

 \item If you are using existing assets (e.g., code, data, models) or curating/releasing new assets, check if you include:
 \begin{enumerate}
   \item Citations of the creator If your work uses existing assets. [Yes. The datasets used in the paper is well-cited.]
   \item The license information of the assets, if applicable. [Not Applicable. ]
   \item New assets either in the supplemental material or as a URL, if applicable. [Yes. We provide the citations to the codes that we used to produce the baselines.]
   \item Information about consent from data providers/curators. [Not Applicable]
   \item Discussion of sensible content if applicable, e.g., personally identifiable information or offensive content. [Not Applicable]
 \end{enumerate}

 \item If you used crowdsourcing or conducted research with human subjects, check if you include:
 \begin{enumerate}
   \item The full text of instructions given to participants and screenshots. [Not Applicable]
   \item Descriptions of potential participant risks, with links to Institutional Review Board (IRB) approvals if applicable. [Not Applicable]
   \item The estimated hourly wage paid to participants and the total amount spent on participant compensation. [Not Applicable]
 \end{enumerate}

 \end{enumerate}


%
%




%

%

\onecolumn
\aistatstitle{Heterogeneous Graph Structure Learning through the Lens of Data-generating Processes: Supplementary Materials}

\appendix

\section{THEORETICAL RESULTS ON HOMOGENEOUS GRAPH STRUCTURE LEARNING}

\subsection{The Data-generating Process for Homogeneous Graphs}
\label{apdx: generative_process}

We wish to derive the probability density function (PDF) for the data-generating process (DGP) on homogeneous graphs. To recall, the DGP for a homogeneous graph can be decomposed as, $P(\mX, \mW, \mY)=P(\mY \mid \mW) P(\mX \mid \mY) P(\mW)$. We are interested in deriving the joint probability of $\mX$ and $\mY$, denoted as $P(\mX, \mY\mid \mW)=P(\mY \mid \mW) P(\mX \mid \mY)$, and then obtain the marginal distribution of $P(\mX \mid \mW)$ by,
\begin{equation}
P(\mX \mid \mW) = \int P(\mY \mid \mW) P(\mX \mid \mY) d\mY,
\end{equation}
which we use as the likelihood term to formalize the maximum a-posteriori (MAP) estimation objective on $\mW$. 

We first show the form of $P(\mY \mid \mW)$. Similar to ~\citet{zhu2003semi}, we choose the node-wise and pair-wise potential as $\varphi_1(v) = \exp(-(d_v+\nu)\|\vy_v\|^2)$ with $\|\cdot\|$ the $L^2$ norm and $\varphi_2(u, v) = \exp(w_{uv}\vy_u^\top\vy_v)$ respectively. This leads to a multi-variate Gaussian on each column of the hidden variable $\mY_{:, k}\in \gR^{|\gV|}$ with zero mean and precision matrix $\Lambda = \mL + \nu \mI$ where $\mI$ is the identity matrix. The overall likelihood function for $\mY$ is then,
\begin{equation}
\begin{aligned}
P(\mY\mid \mW)&=\frac{1}{Z} \prod_{v \in \gV} \varphi_1 \left(\vy_v \right)\prod_{\{v,u\} \in \gE} \varphi_2\left(\vy_v , \vy_u \right)\\
&=\frac{1}{Z} \prod_{v \in \gV} \exp(-(d_v+\nu)\|\vy_v\|^2)\prod_{\{v,u\} \in \gE} \exp(w_{uv}\vy_u^\top\vy_v)\\
&=\frac{1}{Z} \exp \left\{-\sum_{v\in \gV} (d_v+\nu)\|\vy_v\|^2 + \sum_{\{v,u\} \in \gE}w_{uv}\vy_u^\top\vy_v \right\}\\
&=\frac{1}{Z} \exp \left\{-\frac{1}{2}\sum_{k=1}^C(\mY^\top_{:,k}\Lambda\mY_{:,k})\right\}\\
&=\frac{1}{Z} \exp \left\{-\frac{1}{2}\operatorname{tr}(\mY^\top\Lambda\mY)\right\},
\end{aligned}
\end{equation}
where $\operatorname{tr}$ is the trace operator and we simplify the partition function $Z(\mW)$ as $Z$. If each dimension of the hidden variable, $\mY_{:,k}$ is independent to each other, it has a PDF $P(\mY_{:,k}\mid \mW) \propto \exp \left\{-\mY^\top_{:,k}\Lambda\mY_{:,k} \right\}$, which is a multivariate Gaussian with precision matrix $\Lambda$. Please note that the overall likelihood function differs from a simple multivariate Gaussian as $\mY \in \R^{N\times C}$ is a matrix instead of a vector. But we can model the joint probability distribution of two hidden variables $\vy_u$ and $\vy_v$ instead, which is introduced as follows. 

Without loss of interpretability, we denote the concatenation of two hidden variable vectors as $\vy \coloneqq \operatorname{cat}(\vy_v, \vy_u) \in \R^{2C}$, two signal vectors as $\vx \coloneqq \operatorname{cat}(\vx_v, \vx_u) \in \R^{2K}$, and consider $\Lambda_{uv}$ as an entry of the precision matrix. The joint distribution for the hidden variable vectors on a pair of nodes, $\vy_v$ and $\vy_u$ becomes,
\begin{equation}
\label{eq: marginal_y_on_w}
\begin{aligned}
P(\vy \mid \Lambda)&=\frac{1}{Z} \exp \left\{-\frac{1}{2}(\vy^\top\left[\begin{array}{cc}
(d_v+\nu)\mI & \Lambda_{uv}\mI \\
\Lambda_{vu}\mI & (d_v+\nu)\mI
\end{array}\right] \vy)\right\} \\
&=\frac{1}{Z} \exp \left\{-\frac{1}{2}(\vy^\top\left[\begin{array}{cc}
(d_v+\nu)\mI & -\mW_{uv}\mI \\
-\mW_{vu}\mI & (d_v+\nu)\mI
\end{array}\right] \vy)\right\}.
\end{aligned}
\end{equation}

Then, the emission function $P(\mX \mid \mY)$ is considered as another multi-variate Gaussian~\citep{pmlr-v33-liu14} which has a linear relationship, specifically $\vx_v \mid \vy_v  \sim \N (\mV \vy_v, \Sigma_\vx)$. 
We denote $\mV$ as the linear transformation matrix and $\Sigma_x$ the marginal covariance matrix for $\vx_v$, which are shared across all nodes $v\in \gV$. Then, PDF of $P(\mX \mid \mY)$ can be decomposed to nose-wise, namely,
\begin{equation}
P_\mV(\mX \mid \mY)=\prod_{v \in \gV} P_\mV\left(\vx_v \mid \vy_v\right).
\end{equation}
Note that the linear transformation matrix $\mV$ is shared across all nodes $v\in \gV$. Thus, by marginalizing out $\vy_v$, we have the likelihood function for $\vx \equiv \operatorname{cat}(\vx_v, \vx_u)$ as
\begin{equation}
\label{eq: x_given_W}
\begin{aligned}
P(\vx \mid \mW) &= \int P_\mV(\vx\mid \vy)P(\vy \mid \mW)d\vy\\
\text{and }\vx \mid \mW&\sim \N (0, \Sigma_x+\mV^\prime \left[\begin{array}{cc}
(d_v+\nu)\mI & \Lambda_{uv}\mI \\
\Lambda_{vu}\mI & (d_u+\nu)\mI
\end{array}\right] ^{-1} \mV^{\prime\top}).
\end{aligned}
\end{equation}
$\mV^\prime$ is the stack of two matrix $\mV$, i.e. $\mV^\prime = [\mV^\top, \mV^\top]^\top$.

\textit{Proof}: Finding the marginal distribution, $P(\vx \mid \mW)$ given analytical form of $P(\vy \mid \mW)$ and $P(\vx \mid \vy)$ is a problem that arises frequently in Bayesian theorem for Gaussian variables. We are going to use it a few more times so it would be convenient to derive the general results here. To do so, we introduced the following~\cref{lemma: lin_gas_model}.

\begin{lemma} 
\label{lemma: lin_gas_model}
\textbf{Linear Gaussian Model}. ~\citep{ghahramani2001introduction}

Given two vectorized random variables, $\vb$ following a Marginal Gaussian distribution, and $\va$ following a Gaussian distribution conditioned on $\vb$, which has the following form, 
$$
\begin{aligned}
P(\vb) & =\mathcal{N}\left(\vb \mid \boldsymbol{\mu}_1, \Omega_1^{-1}\right) \\
P(\va \mid \vb) & =\mathcal{N}\left(\va \mid \mT \vb+\boldsymbol{\mu}_2, \Omega_2^{-1}\right)
\end{aligned}
$$
where $\mT$ is the linear transformation matrix, $\Omega_1$ and $\Omega_2$ are two precision matrices. The random variable $\va$ has a marginal distribution, 
\begin{equation}
\begin{aligned}
P(\va) = \gN\left(\va\mid \mT\boldsymbol{\mu}_1+\boldsymbol{\mu}_2, \Omega_2^{-1}+\mT \Omega_1^{-1} \mT^{\mathrm{T}} \right)
\end{aligned}
\end{equation}
\end{lemma}

We can simply apply the lemma to~\cref{eq: marginal_y_on_w} and $\vx_v \mid \vy_v  \sim \N (\mV \vy_v, \Sigma_\vx)$, which gives the result in~\cref{eq: x_given_W}.  This ends the proof.

To further simplify the notation, we make the assumption that the marginal noise on $\vx$ is not too strong to conceal the structure information, i.e., $\det(\Sigma_\vx) \ll \det(\Lambda^{-1})$, the PDF function can be obtained, 
\begin{equation}
\begin{aligned}
    P_\mV(\mX\mid \mW) &=\frac{1}{Z(\mW)}  \prod_{v\in \gV} \varphi_1(v) \prod_{\{v,u\} \in \gE}\varphi_2(u,v)
    \\
    \text{with }\varphi_1(v)&= \exp(-(\nu+d_v)\|\mV^{\dagger} \vx_v\|^2) \\
    \text{and }\varphi_2(u,v) &= \exp \left\{ w_{uv} (\vx_u^\top\mV^{\dagger \top}\mV^\dagger\vx_v)\right\},
\end{aligned}
\end{equation}
where $\mV^\dagger$ is the Moore–Penrose pseudo inverse of matrix $\mV$. This is exactly~\cref{eq:likelihood_HMN_signals} thus ends the derivation. 

\subsection{The Optimization Objective on Homogeneous Graph Structure Learning}
\label{apdx: GSL_training_obj}
Here we derive the optimization objective for GSL on homogeneous grahs, which starts from~\cref{eq: map} and targets to~\cref{eq: GSL_format}. According to~\cref{eq:likelihood_HMN_signals}, the parameters contain the weighting matrix $\mW$ (or Laplacian matrix equivalently), transformation matrix $\mV^\dagger$ and we view the scalar $\nu$ considered as hyper-paramter. This yields a MAP problem as introduced in~\cref{eq: map}, we restated as follows,
\begin{equation}
\begin{aligned}
\mW^*, \mV^* &=\argmax_{\mW, \mV} \log P_\mV(\mW \mid \mX) \\&=\argmax_{\mW, \mV} \log P_\mV(\mX \mid \mW) +\log P(\mW) + \log P(\mV).
\end{aligned}
\end{equation}
Current graph structure learning (GSL) algorithms~\citep{DBLP:journals/tsp/DongTFV16, pu2021learning, DBLP:conf/nips/0005YCP19} can be considered as a special case of the above MAP problem when $\mV^{\dagger}$ is rectangular orthogonal (semi-orthogonal) Matrix, so that $\mV^{\dagger \top}\mV^\dagger = \mI$. In such a scenario, the likelihood function of~\cref{eq:likelihood_HMN_signals} reduces to the same form as the Gaussian Graphical Model with,
\begin{equation}
\begin{aligned}
    &P(\mX\mid \mW) =\frac{1}{Z(\mW)}  \prod_{v \in \gV} \exp(-(\nu+d_v)\|\vx_v\|^2) \cdot \prod_{\{v,u\} \in \gE} \exp \left\{ w_{uv} (\vx_u^\top\vx_v)\right\}.
\end{aligned}
\end{equation}
The only parameter left is the weight matrix $\mW = \{w_{ij}\}_{i,j\in\{1:N\}}$ representing the graph structure. Then the problem becomes, $\mW^* =\argmax_{\mW} \log P(\mX \mid \mW) +\log P(\mW )$. One would care about the log-likelihood of the signal $\mX$ for convenient optimization,
\begin{equation}
\begin{aligned}
\log \left[P\left(\mX \mid \mW \right)\right] &= \sum_{v\in \gV} -(d_v + \nu)\|\vx_v\|^2+\sum_{u \neq v} w_{u v} \vx_u^\top\vx_v-\log Z(\mW)\\
&= -\nu \sum_{u=1}^N  \|\vx_u\|^2-\sum_{u,v} \mL_{u v} \vx_u^\top\vx_v-\log Z(\mW),
\end{aligned}
\end{equation}
The step comes from the fact that $\mL = \mD - \mW$. Please note that the summation changed from $u \neq v$ to all the possible $\{u,v\}$ to transit to the Laplacian matrix. Then we obtain the log-likelihood in matrix form,
\begin{equation}
\label{eq: GGM}
\begin{aligned}
\log \left[P\left(\mX \mid \mW \right)\right] 
&= -\nu \sum_{u=1}^N  \|\vx_u\|^2-\sum_{u,v} \mL_{u j} \vx_u^\top\vx_v-\log Z(\mW)\\
&= -\nu \operatorname{tr}(\mX^\top \mX)-\operatorname{tr}(\mX^{\top}\mL \mX)+\log\det(\mL+\nu\mI) + \text{Constant}.
\end{aligned}
\end{equation}
The last step is derived given the only term related to the optimization goal in $Z(\mW)$ is the log-determinant of the precision matrix (negative log-determinant of the covariance matrix). $\nu$ works as the factor balancing the node-wise and pair-wise effects. The pair-wise potentials are scaled by parameter $\mL_{uv}$ which encodes the connection strength.  In graph signal processing, the second term, in the quadratic form of the graph Laplacian, is called the ``smoothness'' of the signal on the graph. As shown in ~\cite{zhou2004regularization}, it can be measured in
terms of pair-wise difference of node signals,
\begin{equation}
\begin{aligned}
    \operatorname{tr} (\mX^\top \mL \mX )=\frac{1}{2} \sum_{\{u,v\}} w_{uv}\|\vx_u-\vx_v\|^2,
\end{aligned}
\end{equation}
given that $\sum_v w_{uv} = \mD_{uu}$. Intuitively this term suggests, if two signal vectors $x_u$ and $x_v$ from a smooth set reside on two well-connected nodes with large $w_{uv}$, they are expected to have a small dissimilarity.

Plug~\cref{eq: GGM} into the training objective will give the same form as Graphical Lasso~\citep{DBLP:journals/jmlr/BanerjeeGd08}. However, the computationally demanding log-determinant term makes the problem difficult to solve~\citep{DBLP:conf/aistats/Kalofolias16}. Thus, ~\citet{DBLP:journals/jstsp/EgilmezPO17} consider a relaxed optimization on the lower bound instead. 
\begin{lemma}
\label{lemma: log_deter_relax}
\textbf{The Log-determinant Lower Bound.}

The optimization objective, 
$$\argmin_\Lambda - \sum_v\log \left(\operatorname{det}(\Lambda)_v\right),$$
can be relaxed through $$\argmin_\Lambda - \sum_v\log \left(\operatorname{diag}(\Lambda)_v\right).$$
\end{lemma}

\textit{Proof.} By using Hadamard's inequality, we get,$
\operatorname{det}(\Lambda) \leq \prod_v \Lambda_{v v}$. So that,
\begin{equation}
\log (\operatorname{det}(\Lambda)) \leq \log \left(\prod_v \Lambda_{v v}\right)=\sum_v \log \left(\Lambda_{v v}\right).
\end{equation}
After rearranging the terms, we derive the following lower bound that $- \sum_v\log \left(\operatorname{diag}(\Lambda)_v\right) \leq-\log \operatorname{det}(\Lambda)$, which ends our proof.

Using~\cref{lemma: log_deter_relax} and substituting the log-determinant with~\cref{eq: GGM}, we can further derive the optimization objective as,
\begin{equation}
\label{eq: GSP_DGP_apdx}
\begin{aligned}
    &\argmax_{\mW} \log P(\mX \mid \mW) +\log P(\mW )\\
    \rightarrow &\argmax_{\mL} -\nu \operatorname{tr}((\mX^\top \mX)-\operatorname{tr}(\mX^{\top}\mL \mX)+ \mathbf{1}^\top\log \left(\operatorname{diag}(\mL+\nu\mI)_v\right) +\log P(\mW)\\
    =& \argmin_\mW \underbrace{\sum_{ij} w_{ij} \|\vx_i -\vx_j\|^2}_{S(\mX, \mW)} \underbrace{- \mathbf{1}^\top\log \left(\mW \cdot \mathbf{1}\right) -\log P(\mW)+ \nu \|\mX\|^2|}_{\Omega(\mW)}\\
    \equiv& \argmin_\mW S(\mX, \mW) + \Omega(\mW),
\end{aligned}
\end{equation}
The training objective consists of graph-signal fidelity term $S(\mX, \mW)$ and a structural regularizer $\Omega(\mW)$, which is exactly our training objective in ~\cref{eq: GSL_format} if we absorb the $L^2$ regularizer into the negative log-prior.

\textbf{Remark.} To transit from the precision matrix MAP estimation through the parameterization of hidden Markov Networks~\cite{ghahramani2001introduction, yuan2007model} to graph structure learning problem, three important assumptions are made: 1) Each feature dimension in $\mX$ contributes equally to the structure estimation, i.e,$\mV^\dagger$ is semi-orthogonal; 2) the variance of $\vx$, $\Sigma_x$ is considered to be relatively small to the feature generation covariance across all the nodes, so the graph structure dominates the conditional covariance matrix of $\mX$, i.e. $\det(\Sigma_\vx) \ll \det(\Lambda^{-1})$, and 3) The log-determinant of the partition function is relaxed by the lower-bound in the optimization. 

\section{THEORETICAL RESULTS ON HETEROGENEOUS GRAPH STRUCTURE LEARNING}

\subsection{The Data-generating Process for Heterogeneous Graphs}
\label{subsec: DGM_HG}
Similarly, we want to derive the likelihood term $P(\{\vx_v\} \mid \tW, \{\mB_r\})$ so that we can formalize the MAP estimation objective. We first model the joint PDF of $\{\vx_v\}$ and $\{\vy_v\}$ through the following form,
\begin{equation}
\begin{aligned}
    &P(\{\vx_v\}, \{\vy_v\}\mid \tW, \{\mB_r\}) \\
    =&P(\{\vy_v\}\mid \tW, \{\mB_r\}) \prod_v P\left(\vx_v \mid \vy_v \right) \\
    = &\frac{1}{Z}\prod_v  \varphi_1\left(v \right) \prod_{u\neq v,r} \varphi_r \left(u, v\right)\prod_v P\left(\vx_v \mid \vy_v \right)
\end{aligned}
\end{equation}
where the node-wise and edge-wise potential functions are respectively $\varphi_1(v) = \exp(-(d_v+\nu)\|\vy_v\|^2)$ and $\varphi_r(u, v) = \exp(w_{uvr}\vy_u^\top\mB_r\vy_v)$. $Z$ is the partition function and $d_v$ is the degree of node v, $d_v = \sum_{ru} w_{uvr}$. Multiple distribution types belonging to the exponential family satisfy the form in~\cref{eq:joint_likelihood}. For simplicity, we view the hidden variables as the continuous embedding of labels and assume $P(\vy_u, \vy_v)$ is a joint Gaussian that has, 
\begin{equation}
\operatorname{cat}
([\vy_u, \vy_v])  \sim \gN\left(
\left[\begin{array}{cc}
\mu_{\vy_v} \\
\mu_{\vy_u}
\end{array}\right],
\left[\begin{array}{cc}
(d_u+\nu)\mI & -w_{vur}\mB_r \\
-w_{vur}\mB_r^\top & (d_v+\nu)\mI
\end{array}\right]^{-1}\right),
\end{equation}
where $\operatorname{cat}$ is the concatenation operator. It is noted that the off-diagonal entries in the precision matrix are no longer a diagonal matrix but instead controlled by $\mB_r$, which suggests the intertwined nature between different dimensions of the hidden variable vector $\{\vy_v\}$. Given all the above necessary notations, we can now get the generation density of $\{\vy_v\}$ in heterogeneous graphs as follows,
\begin{equation}
P(\{\vy_v\}\mid \tW, \{\mB_r\}) \propto \exp{}(- \sum_{v}
(d_v+\nu)\|\vy_v\|^2 + \sum_{u \neq v, r} w_{vur} \vy_u^\top \mB_r \vy_v),
\end{equation}
For now we use ``proportional to'' and ignore the partition function to simplify the notation. The partition function will be handled with a lower bound in the optimization objective using~\cref{lemma: log_deter_relax}. 

From our assumption, the conditional distribution of $P(\vx_v \mid \vy_v)$ has a mean that is linearly transformed from $\vy_v$, and a covariance that is independent of $\vy_v$, which gives\footnote{From now on, all the precision matrices are termed as $\Lambda$ and covariance matrix as $\Sigma$.},
\begin{equation}
P(\vx_v \mid \vy_v) = \N\left(\vx_v \mid \mV_{\phi(v)} (\vy_v-\mu_{\vy_v}), \Sigma_{\vx_v} \right)
\end{equation}
where we assume that the linear transformation matrix $\mV_{\phi(v)}\in \R^{K\times |C_{\phi(u)}|}$ is determined by the node type $\phi(v)$ and simply consider the covariance matrix $\Sigma_{\vx_v} = \sigma^2\mI$, which gives
~\cref{eq: emission_func}. Thus, the joint distribution $\vy_v$ and $\vx_v$ also follow a Gaussian distribution, which gives, 
\begin{equation}
\left[\begin{array}{c}
\vx_v \\
\vy_v
\end{array}\right]
\sim \N\left(
\left[\begin{array}{cc}
0 \\
\mu_{\vy_v}
\end{array}\right],
\left[\begin{array}{cc}
\Sigma_{\vx_v} + \mV_{\phi(v)}\Sigma_{\vy_v} \mV_{\phi(v)}^\top& -\Sigma_{\vy_v} \mV_{\phi(v)}^\top \\
 -\mV_{\phi(v)}\Sigma_{\vy_v} & \Sigma_{\vy_v}
\end{array}\right]\right).
\end{equation}
Until now we have derive the $P(\vx_v \mid \vy_v)$ and $P(\vy_u, \vy_v \mid \lambda_{uvr})$ where for simplicity the parameters are denoted as $\Theta = \left\{\tW, \{\mB_r\}\right\}$. Recall that $\vx \coloneqq \operatorname{cat}(\vx_v, \vx_u)$ and $\vy \coloneqq \operatorname{cat}(\vy_v, \vy_u)$. We can apply~\cref{lemma: lin_gas_model} and obtain the conditional distribution of $P(\vx \mid \Theta)$ also being a Gaussian with the following joint likelihood function,
\begin{equation}
\label{eq:pairwise_likelihood}
\begin{aligned}
    &P(\vx \mid \Theta) \propto \exp  \left\{ -\frac{1}{2}(\vx^\top \Lambda_{\vx \mid \Theta} \vx)\right\},
\end{aligned}
\end{equation}
where the mean $\mu_{x\mid\Theta}=0$ and the precision matrix, 
\begin{equation}
\begin{aligned}
    &\Lambda_{\vx \mid \Theta} =(\Sigma_{\vx} + \mV\Sigma_{\vy \mid \Theta} \mV^\top)^{-1}
\end{aligned}
\end{equation}
with $\Sigma_{\vy\mid\Theta} = \Lambda_{\vx\mid\Theta}^{-1}$ the covariance matrix. $\mV = \left[\mV_{\phi(v)}^\top, \mV_{\phi(u)}^\top\right]^\top$. If we consider the variance of $\vx$ is relatively small compared to the graph emission matrix $\sigma_{\vy}$, i.e. $\det(\Sigma_\vx) \ll \det(\mV\Sigma_{\vy \mid \Theta} \mV^\top)$, we obtain,
\begin{equation}
\begin{aligned}
&\Lambda_{\vx \mid \Theta} = ( \Sigma_{\vx} + \mV\Sigma_{\vy \mid \Theta} \mV^\top)^{-1}\approx\\
&\left(\left[\begin{array}{c}
\mU_{\phi(u)}\\
\mU_{\phi(v)}
\end{array}\right]^\top
\left[\begin{array}{cc}
(d_u+\nu)\mI & -w_{vur}\mB_r \\
-w_{vur}\mB_r^\top & (d_v+\nu)\mI
\end{array}\right]
\left[\begin{array}{c}
\mU_{\phi(u)} \\
\mU_{\phi(v)}
\end{array}\right]\right)
\end{aligned}
\end{equation}
where $\mU_{\phi(u)} \in \R^{|C_{\phi(u)}|\times K}$ is the Moore–Penrose Pseudo inverse of matrix $\mV_{\phi(u)}$ which satisfies $\mV_{\phi(u)}\mU_{\phi(u)}\mV_{\phi(u)} = \mV_{\phi(u)}$ and $\mU_{\phi(u)}\mV_{\phi(u)}\mU_{\phi(u)} = \mU_{\phi(u)}$. For the details of such a pseduo inverse method, we refer to~\cite{pseduo_inverse} for a tutorial. With such a derivation, we can get
\begin{equation}
\label{eq:precision}
\begin{aligned}
    -\Lambda_{\vx \mid \Theta} &= 
    -(d_u+\nu)\mU_{\phi(u)}^\top\mU_{\phi(u)} - (d_v+\nu)\mU_{\phi(v)}^\top\mU_{\phi(v)} 
    + 2w_{uvr} \mU_{\phi(u)}^\top \mB_r \mU_{\phi(v)}.
\end{aligned}
\end{equation}

We then combine~\cref{eq:precision} and~\cref{eq:pairwise_likelihood} to obtain the overall likelihood for the whole graph signals, 
\begin{equation}
\begin{aligned}
    &P(\{\vx_v\}\mid \tW, \{\mB_r\}) \propto \prod \exp  \left\{ -\frac{1}{2}(\vx^\top \Sigma^{-1}_{\vx \mid \Theta} \vx)\right\}\\
    & = \prod_v \exp(-(\nu+d_v)\|\mU_{\phi_v} \vx_v\|_\Frm^2)\cdot \prod_{u,v,r} \exp \left\{ w_{uvr}(\vx_u^\top\mU_{\phi(u)}^\top \mB_r \mU_{\phi(v)}\vx_v)\right\}
\end{aligned}
\end{equation}
It is worth pointing out that the matrix $\mU_{\phi(u)}^\top \mB_r \mU_{\phi(v)}$ is positive semi-definite if the signals on $u$ and $v$ lies in the same space. This is not always true but in this paper we make this assumption.

\subsection{The Optimization Objective for Heterogeneous Graph Structure Learning}
\label{apdx: training_obj_hgsl}
We first consider the likelihood function:
\begin{equation}
\begin{aligned}
    &P(\{\vx_v\}\mid \tW, \{\mB_r\}) \propto \prod_v \varphi_1(v) \prod_{u\neq v,r} \varphi_r(u,v) \\
    & \text{with } \varphi_1(v) = \exp(-(\nu+d_v)\|\mU_{\phi_{(v)}} \vx_v\|^2) \text{ and }\varphi_r(u,v) = \exp \left\{ w_{uvr} (\vx_u^\top\mU_{\phi(u)}^\top \mB_r \mU_{\phi(v)}\vx_v)\right\}.
\end{aligned}
\end{equation}

\begin{lemma}
If $\mP \in \gR^{n\times n}$ is positive-semidefinite, there always exits a matrix $\mQ$ that has the same rank as $\mP$, so that $\mQ^\top\mQ= \mP$.
\end{lemma}

Recall that the type of edge can help determine the node types connected (reversely is not possible). With the lemma, we can use a relation-specific matrix $\mM_r \in \gR^{K\times K^\prime}$ to replace $\vx_u^\top \mM_r\mM_r^\top \vx_v = \vx_u^\top\mU_{\phi(u)}^\top \mB_r \mU_{\phi(v)}\vx_v$ for each type of relation $M_r$. So the log-potential becomes,
\begin{equation}
\label{eq: HSBM_opt_Y}
\begin{aligned}
 &\sum_{u \neq v, r} w_{uvr} (\vx_u^\top\mM_r \mM_r^\top \vx_v) - \sum_v (\nu+d_v)\|\mU_{\phi(v)} \vx_v\|^2
  \end{aligned}
 \end{equation}
It is noted that the intra-type node connection (nodes connected within the same type) can be viewed as a special relation type. In such a scenario, the connectivity matrix $\mB_r$ is a positive symmetric matrix, so it is possible to find a matrix $\mM_r$ such that $\sum_v \|\vx_v^\top\mU_{\phi(v)}^\top\mB_r \mU_{\phi(v)} \vx_v\|^2 = \sum_v \|\mM_r \vx_v\|^2$. It is also easier for us to unify the notation. This leads to the following simplification,
 \begin{equation}
 \begin{aligned}
  &\sum_{u \neq v, r} w_{uvr} (\vx_u^\top\mM_r \mM_r^\top \vx_v) - \sum_v (\nu+d_v)\|\mU_{\phi(v)} \vx_v\|^2\\
  =&\sum_{u \neq v, r} w_{uvr} (\vx_u^\top\mM_r \mM_r^\top \vx_v) - \sum_r \sum_v (\nu+d_v)\|\mM_r^\top \vx_v\|^2\\
 =&-\frac{1}{2}\sum_{u, v, r}  w_{uvr}\|\mM_r^\top\vx_{u}-\mM_r^\top\vx_{v}\|^2- \nu\sum_r \sum_v\|\mM_r^\top \vx_v\|^2
 \end{aligned}
\end{equation}

The last step comes from the fact that $\sum_{r,v} w_{uvr} = d_v$ and the submission subscript changes from $u \neq v$ to all possible combinations $u,v$. If we take $\nu \rightarrow 0$, the only term left is 
$$-\frac{1}{2}\sum_{u, v, r}  w_{uvr}\|\mM_r^\top\vx_{u}-\mM_r^\top\vx_{v}\|^2.$$ 
We next consider a low-rank approximation such that,
$$\mM_r \mM_r^\top \approx \mU_{\phi(u)}^\top \mB_r \mU_{\phi(v)},$$
which would further simplify the parameter setting. Our empirical study shows that a rank-1 relation embedding $\ve_r \in \gR^K$ is sufficient in the approximation. Thus, the training objective leads to,
\begin{equation}
\begin{aligned}
\argmin_{\tW, \mE} &\sum_{v\leq u, r} w_{vur}\| \ve_r^\top\cdot(\vx_v- \vx_u)\|^2 + \Omega(\tW) + \Omega (\mE)
\end{aligned}
\end{equation}
which is exactly~\cref{eq:training_obj}. It is noted that we absorb the partition function into the regularizer. And such a term can be relaxed by the log-barrier on the degree matrix, similar to what we did in~\cref{eq: GSP_DGP_apdx} using~\cref{lemma: log_deter_relax}.

\subsection{The relation embedding update step in HGSL optimization}
\label{apex: der_relation_emb}

Here we gave a detailed analysis of algorithm 1 when updating relation embeddings, especially when we transit from~\cref{eq: relation_emb_update} to~\cref{eq: FR}. We first repeat the training objective here, 
\begin{equation}
\begin{aligned}
\argmin_{\{\ve_r\}} \quad &\sum_{vur} w_{vur}\| \ve_r^\top \cdot (\vx_v- \vx_u)\|^2 + \lambda_1 \|\mE\|^2 + \lambda_2 \|\mE\|_1.
\end{aligned}
\end{equation}

As we show in~\cref{apdx: training_obj_hgsl}, the training objective is equivalent to the following:
\begin{equation}
\begin{aligned}
\argmin_{\{\ve_r\}} \quad &-\sum_{vur} w_{vur}  (\vx_u^\top \ve_r \ve_r^\top \vx_v) + \lambda_1 \sum_r \ve_r^\top \cdot \ve_r + \lambda_2 \sum_r \|\ve_r\|_1.
\end{aligned}
\end{equation}
It is possible to disentangle through the relation types and get the sub-problem as,
\begin{equation}
\begin{aligned}
\argmin_{\ve_r} \quad &-\sum_{vu} w_{vur}  (\vx_u^\top \ve_r \ve_r^\top \vx_v) + \lambda_1  \ve_r^\top \cdot \ve_r + \lambda_2  \|\ve_r\|_1.
\end{aligned}
\end{equation}
Denote $\mP_r=\sum_{u, v} w_{u v r} \vx_u \vx_v^{\top}$, and notice that $\vx_v^{\top} \ve_r \cdot \ve_r^{\top} \vx_u=\ve_r^{\top} \vx_u \vx_v^{\top} \ve_r$. We get $
-\sum_{u, v} w_{u v r}\left(\vx_v^{\top} \ve_r\right)\left(\ve_r^{\top} \vx_u\right)=-\ve_r^{\top} \mP_r \ve_r$ and the objective function simplifies to:
\begin{equation}
\begin{aligned}
\gO\left(\ve_r\right) &=-\ve_r^{\top} \mP_r \ve_r+\lambda_1 \ve_r^{\top} \ve_r+\lambda_2\left\|\ve_r\right\|_1\\
&=\ve_r^{\top}(\lambda_1 \mI - \mP_r) \ve_r+\lambda_2\left\|\ve_r\right\|_1
\end{aligned}
\end{equation}

We wish to ensure the $\ve_r\geq 0$ to be non-negative so that we set up the Lagrangian with Non-negativity Constraints using Lagrange multipliers $\gamma \geq 0$, and obtain,
\begin{equation}
L\left(\ve_r, \gamma\right)=\ve_r^{\top} (\lambda_1 \mI - \mP_r) \ve_r+\lambda_2 \ve_r^{\top} 1-\gamma^{\top} \ve_r
\end{equation}
Setting the derivative to zero, we get
$\frac{\partial L}{\partial \ve_r}=2 (\lambda_1 \mI - \mP_r) \ve_r+\lambda_2-\gamma=0$ and Complementary Slackness:
$\gamma_i \ve_{r i}=0, \forall i$. Solve for $e_r >0$, we get $\lambda_i=0$, so the stationarity condition becomes:
\begin{equation}
\frac{\partial L}{\partial \ve_r}=2 (\lambda_1 \mI - \mP_r) \ve_r+\lambda_2 \mathbf{1}=0
\end{equation}
And this leads to the analytical solution, with
\begin{equation}
\ve_r = \frac{\lambda_2}{2}(\mP_r - \lambda_1\mI)^{-1}\mathbf{1}
\end{equation}

We assume $\|\frac{1}{\lambda_1}\mP_r - \mI\|<1$, thus we can use first-order Neumann Series Expansion and yield:
\begin{equation}
\ve_r \approx \frac{\lambda_2}{2\lambda_1}(\frac{1}{\lambda_1}\mP_r-2\mI)\mathbf{1}
\end{equation}
replace $\mP_r=\sum_{u, v} w_{u v r} \vx_u \vx_v^{\top}$, we obtain:
\begin{equation}
\ve_r = \frac{\lambda_2}{2\lambda_1}(\frac{1}{\lambda_1}\sum_{u, v} w_{u v r} \vx_u \vx_v^{\top} - 2\mI)\mathbf{1}
\end{equation}
Looking into each dimension will yield~\cref{eq: FR}, which has an entry-wise solution as:
\begin{equation}
\ve_{r, k}^{t+1} = \frac{\lambda_2}{2\lambda_1}(\frac{1}{\lambda_1}\sum_{\{v, u, r\} \in \E^\prime_r} w_{vur}^t~ \vx_{v,k} \cdot \vx_{u,k}-2)
\end{equation}
In practice, we rewrite $\alpha = \frac{\lambda_2}{2\lambda_1^2}$ and $\beta = \frac{\lambda_2}{\lambda_1}$, and the solution becomes 
\begin{equation}
\ve_{r, k}^{t+1} = \alpha\sum_{\{v, u, r\} \in \E^\prime_r} w_{vur}^t~ \vx_{v,k} \cdot \vx_{u,k} -\beta
\end{equation}
We do a hyperparameter search for the $\alpha$ and $\beta$.

\section{THE EXPRESSIVENESS OF H2MN}
\label{apdx: expressiveness}

Designing a statistical model that effectively captures the heterogeneity of node and edge types is a central challenge in extending Graph Structure Learning (GSL) algorithms to Heterogeneous Graph Structure Learning (HGSL). This challenge stems from three key aspects. First, the complex dependencies among various components, including graph structures with multiple relation types, node features, and node labels, necessitate a model that can capture these multifaceted dependencies. Second, there exists an intertwined dependency among edges with different relation types, and modeling this interdependency is crucial. Finally, the node feature generation needs to consider the influence on different node labels and relation types.

We first answer the question about why the dependencies exist and how our model design captures the dependency among edges with different relation types.

\begin{propensity}
    The probabilities of edges in the same graph having specific relation types are not independent under the assumption of H2MN.
\end{propensity}

To illustrate, we consider a simple path in the graph $\mathcal{P}: v_1 \stackrel{e_1}{\longleftrightarrow} v_2 \stackrel{e_2}{\longleftrightarrow} v_3$, where $e_i$ denote edges, $e_i=r$ indicates the edge has relation type $r$. We consider the probability of $e_2$ having type $r$ given $e_1$ having $r^\prime$, $P(e_2=r\mid e_1 =r^\prime)$, with $r$ and $r^\prime$ two different relation types. Considering the variables follow the Markov property, we can derive that $P\left(e_2 = r \mid e_1 = r^\prime\right)=\sum_{v_2} P\left(e_2 =r\mid v_2\right) \cdot P\left(v_2 \mid e_1 = r^\prime\right)$. There is no way we can further eliminate the dependency on $e_1$, which means $P\left(e_2 = r \mid e_1 = r^\prime\right) \neq P\left(e_2 = r\right)$. So the random variables $e_1$ and $e_2$ are not independent. This necessitates joint statistical modeling of the edges with various relation types rather than treating them independently.

\paragraph{The correlation modeling of H2MN}

We wish to emphasize that, even though we consider the pair-wise HMN to approximate the overall joint probability distribution function, the approximation preserves the dependency among edges with diverse relation types. To illustrate, we consider the example of a path in the graph $\mathcal{P}: v_1 \stackrel{e_1}{\longleftrightarrow} v_2 \stackrel{e_2}{\longleftrightarrow} v_3$ again, and factorize the joint PDF as (omitting the node-wise potential $\varphi(v)$ for simplicity), $$P(\mathcal{P})= \frac{1}{Z} \varphi_{r}(v_1, v_2) \varphi_{r^\prime}(v_2, v_3)$$ where $r$ is the type for $e_1$ and $r^\prime$ is the type for $e_2$, and $Z$ is the partition function. And then we can obtain $$P(e_2 \mid e_1) =\frac{P(e_2, e_1)}{P(e_1)} = \frac{\sum_{v_1, v_2, v_3} \varphi_{r}(v_1, v_2) \cdot \varphi_{r^\prime} (v_2, v_3) }{\sum_{v_1, v_2, v_3, e_2} \varphi_{r}\left(v_1, v_2\right) \cdot \varphi_{r^\prime} \left(v_2, v_3\right)}$$ In this expression, the influence of $e_1$ is retained in the numerator of the fractional above, demonstrating that our model inherently captures the interdependencies among edges with different relation types. The correlation is implicitly modeled within the parameterization of the HMN.

\section{EXTENSION OF THE ALGORITHM}

\subsection{Learning the heterogeneous graphs when the node labels are observable.}
\label{apdx: HGSL_labels}

When labels are observable, such as in IMDB and ACM dataset~\cite{DBLP:conf/www/0004ZMK20}, we wish to learn $\tW$ and $\{\mB_r\}$ directly from the labels $\{\vy_v\}_{v \in \gV}$. To this end, we consider the HGSL problem as a MAP estimation problem parameterized by~\cref{eq: generation_distribution} and define $\Omega(\cdot)$ as the negative log prior, which gives,
\begin{equation}
\begin{aligned}
&\argmax_{\tW, \{\mB_r\}} \log P(\tW, \{\mB_r\} \mid  \{\vy_v\})\\
=&\argmax_{\tW, \{\mB_r\}} \log P(\{\vy_v\} \mid \tW, \{\mB_r\} ) - \Omega(\tW) - \Omega(\{\mB_r\} ).
\end{aligned}
\end{equation}
The log-likelihood follows~\cref{eq: generation_distribution}, thus we derive the training objective as,
\begin{equation}
\begin{aligned}
\argmax_{\tW, \{\mB_r\}} &\sum_{u\neq v, r} (\vy_u^\top\mB_r\vy_v)\cdot w_{uvr} - \sum_{v}
(d_v+\nu)\|\vy_v\|_\Frm^2 -\Omega(\tW) - \Omega(\{\mB_r\} ).
\end{aligned}
\end{equation}
Given that $d_v = \sum_{r,u}w_{vur}$, we can further derive first two terms as,
\begin{equation}
\label{eq:likelihood}
\begin{aligned}
 &- \sum_{v}
(d_v+\nu)\|\vy_v\|^2 + \sum_{\{u\neq v, r\} \in \E} w_{vur} \vy_u^\top \mB_r \vy_v\\
=& - \nu\sum_{v}
\|\vy_v\|^2 + \sum_{u \neq v, r} w_{vur} \vy_u^\top \mB_r \vy_v - \sum_{v} d_v
\|\vy_v\|^2\\
= &-\frac{1}{2}\sum_{u, v, r} w_{uvr} \|\vct (\mB_r)\odot(\vy_u\otimes \mathbf{1}_v - \vy_v\otimes \mathbf{1}_u)\|^2 - \nu\sum_{v}
\|\vy_v\|^2
\end{aligned}
\end{equation}

Here $\vct$ is the vectorization function that flattens the matrix into a long vector, $\otimes$ is the Kronecker product, $\odot$ is the element-wise product, and $\mathbf{1}_v$ is the all-1 vector that has the same size as $y_v$. The node-wise effects solely related to $\vy_v$ do not affect the optimization over $\tW$ and $\{\mB_r\}$ so they can be omitted. Thus, we get, 
\begin{equation}
\label{eq: HGSL_with_label}
\argmin_{\tW, \{\mB_r\}}\sum_{u, v, r} \vct (\mB_r) \|\vy_u\otimes \mathbf{1}_v - \vy_v\otimes \mathbf{1}_u\|^2_2 \cdot w_{uvr}+ \Omega(\tW) + \Omega(\{\mB_r\} ).
\end{equation}
This is the training objective we are going to use if we want to learn the graph structure from node labels $\{\vy_v\}$.

\section{ALGORITHM ANALYSIS: INTUITION AND THEORY}
\label{sec: SBM_HG}

In the vanilla GSL problem, it is widely acknowledged that ``on a homophily graph, minimizing the smoothness could lead to a meaningful graph structure'', just like what we did in~\cref{sec: DGP_GSL_connection}. In~\cref{apdx: homo_gsl_dgp_proof}, we gave the first attempt to understand how the homophily concepts and smoothness assumption would impact the GSL algorithms from the perspective of the data-generating process. In more complex graphs, e.g. heterogeneous graphs, the definition of smoothness and homophily is never properly stated. We mange to link the smoothness in heterogeneous graphs with the DGP proposed in~\cref{subsec:GSGS}. Next, in~\cref{apdx: homophily_proof}, We move to heterogeneous cases to see how we can interpret the homophily in heterogeneous graphs through the H2MN, and derive the condition that our algorithm, proposed in~\cref{alg: main}, will converge to a meaningful solution.

\subsection{Understanding the Graph-data Fidelity Term through Generalized Smoothness}
\label{subsec:GSGS}

In the HGSL training objective as in~\cref{eq:training_obj}, the first term in the training objective is the graph-data fidelity term, 
\begin{equation}
\label{eq: generalized_smooth}
\begin{aligned}
    S (\mX, \mE, \tW) &= \sum_{\{v, u, r\} \in \E} w_{vur}\|\ve_r\circ(\vx_v- \vx_u)\|^2 
    \\&=\langle\tW, \| (\mX\otimes\mathbf{1}- \mathbf{1}\otimes\mX) \otimes \mE\|_F^2\rangle,
\end{aligned}
\end{equation}
where $\circ$ is the element-wise product and $\|\cdot \|_F$ is the Frobenius norm along the last dimension which collapses the tensor from dimension 4 to 3. $\mX = [\vx_1, ..., \vx_{|\mathcal{V}|}]^\mT \in \mathbb{R}^{|\mathcal{V}| \times K}$, $\mE = [\ve_1, \ldots, \ve_{|\mathcal{R}|}]^\mT \in \mathbb{R}^{|\mathcal{R}| \times K }$, and $\mathbf{1} \in \mathbb{R}^{|\mathcal{V}| \times K}$ is an all-one matrix. 
$\langle \cdot, \cdot\rangle$ and $\otimes$ are the tensor inner product and outer product, respectively. 

\begin{figure}
    \centering    \includegraphics[width=0.6\textwidth]{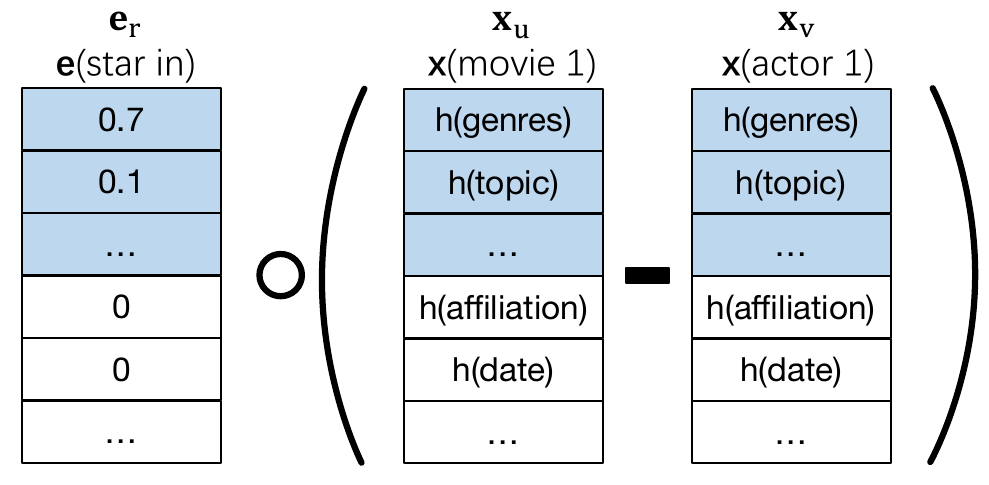}
\caption{Visualization of the generalized smoothness.}
\label{fig: gen_smooth}
\end{figure}

The graph-data fidelity term in~\cref{eq: GSL_format} measures dimension-wise smoothness via $\sum_{u,v} \mW_{u,v}\|\vx_v - \vx_u\|^2$, and sum over different dimensions without any weighting. Similarly in HGSL, if the relation embedding $\ve_r$ is normalized and constrained to be positive, the graph-data fidelity term can be interpreted as a weighted smoothness, as illustrated in~\cref{fig: gen_smooth}. The term can be considered as a \emph{reweighted smoothness} scheme that first measures dimension-wise smoothness, and then integrates it by emphasizing specific signal dimensions according to $r$. Using the movie review dataset again as an example, while determining whether a `star in'-typed edge should be formed between two actor nodes, the model should put larger weights on the signal dimensions that represent the `genres', but weigh less irrelevant ones such as `company affiliation' or `date'.

Different from the original smoothness, the contribution of each signal dimension to the overall smoothness is reweighted by $\ve_r$. Intuitively, a heterogeneous graph is thought to be smooth if strongly connected nodes (with larger $w_{vur}$) have similar signal values in the dimensions emphasized by $\ve_r$ for the relation $r$. 



\subsection{The Optimization Conditions for Graph Structure Learning in Homogeneous Graphs}
\label{apdx: homo_gsl_dgp_proof}

We will first derive the optimization condition based on understanding the homophily property in homogeneous graphs, and then we slowly move toward the heterogeneous graph cases.

We start by introducing the homophily ratio. The concept of homophily in networks was first proposed in~\cite{doi:10.1146/annurev.soc.27.1.415}, suggesting that a connection between similar entities occurs at a higher probability than among dissimilar entities. Statistically, this suggests that we have a higher probability of observing an edge between two nodes belonging to the same communities. Such a concept gives,
\begin{equation}
P(\mW_{uv} =1 \mid y_u = y_v) > P(\mW_{uv}=1 \mid y_u \neq y_v),
\end{equation}
where $y_u$ and $y_v$ are the labels for nodes $u$ and $v$. For simplicity, we denote $P(\mW_{uv} =1 \mid y_u = y_v)=p$ and $P(\mW_{uv}=1 \mid y_u \neq y_v)=q$. And this definition gives the homophily ratio in homogeneous graphs, as follows. 

\paragraph{Expected Homophily Ratio. }
$$\text{HR in homogeneous graphs} = \frac{P(\mW_{uv} =1 \mid y_u = y_v) }{P(\mW_{uv} =1 \mid y_u = y_v) +P(\mW_{uv}=1 \mid y_u \neq y_v)} = \frac{p}{p+q}.$$

\paragraph{Empirical Homophily Ratio.} While a more practical measurement would be the empirical homophily ratio (EHR), defined as,  
\begin{equation}
\text{EHR}\left(\mathcal{G}\right)=\frac{\sum_{\{u, v\} \in \mathcal{E}} \mathbb{I}\left(y_u=y_v\right)}{\left|\mathcal{E}\right|}.
\end{equation}
Which in expectation gives expected homophily ratio.

We are interested in how the homophily parameters $p$ and $q$ are related to the GSL optimization solution. To this end, we first consider the case when $\mB$ is a $2 \times 2$ matrix encoding only the probabilities of having an edge and no edge among nodes with the same label, p, and q, where,
\begin{equation}
    \mB = \left[\begin{array}{cc}
p & q \\
q & p
\end{array}\right].
\end{equation}
Usually, the parameterization considers the average degrees $\bar{d}$ and the total number of nodes $N$, so that $p = \bar{d}\hat{p}/N$ and $q = \bar{d}\hat{q}/N$ with $d \ll N$, $\hat{p}+\hat{q} = 1$. We consider the GSL problem when labels are given for simplicity and a $L^1$ regularizer on $\mW$, which yields the following optimization objective:
\begin{equation}
\label{eq: understand_homophily}
\begin{aligned}
&\argmax_\mW\log P(\mY\mid \mW)+ \log P(\mW) \\
    &= \argmax_\mW\sum_{i\leq j}\vy_i^\top \B\vy_j\cdot\mW_{ij} - \beta \|\mW\|_1\\
    &= \argmax_\mW\sum_{\{i,j\}: y(i)= y(j)} p \mW_{ij} + \sum_{\{i,j\}: y(i) \neq y(j)}q \mW_{ij} - \beta \|\mW\|_1\\
    &= \argmax_\mW \sum_{\{i,j\}: y(i)= y(j)} (p-q) \mW_{ij} + \sum_{\{i,j\}: y(i) = y(j)}q \mW_{ij} + \sum_{\{i,j\}: y(i) \neq y(j)}q \mW_{ij} - \beta \|\mW\|_1\\
    &= \argmax_\mW \quad (p-q) \sum_{\{i,j\}: y(i)= y(j)} \mW_{ij} + (q-\beta) \|\mW\|_1.
\end{aligned}
\end{equation} 
In such a situation, the training objective can be decomposed into two parts, and the optimization solution becomes a trade-off game between the two components. The first part suggests that the algorithm should put edges between nodes with the same label, and the second encourages the sparsity of the solution. To avoid trivial solution (i.e. simply maximizing $\|\W\|_1$ or setting $\|\W\|_1 = 0$) for the problem, we need to have $p-q >0$ and $q-\beta <0$, i.e the graph is homophily with $p>q$. 

This suggests that, only on graphs with high homophily edge generation probability $p$, the GSL algorithm can achieve a meaningful solution. Thus, to conduct the algorithm analysis in the HGSL algorithm, we can similarly define the homophily and derive the conditions.

\subsection{Algorithm Analysis for Heterogenous Graph Structure Learning}
\label{apdx: homophily_proof}

Similar to the homogeneous graphs, we wish to understand what setting of $\B_r, r\in \mathcal{R}$ could give a guarantee on the optimization problem of heterogeneous graph structure learning. We again consider the optimization problem when labels $\{\vy_v\}$ are observable, as introduced in~\cref{eq: HGSL_with_label} but ignore the regularizer for now. Considering all possible relation types as $r \in \gR_{\phi(u)\phi(v)}$ and denoting an element of possible node label combinations within two node types as $ (p, q)\in \mathcal{Y}_{\phi(u)}\times\mathcal{Y}_{\phi(v)}$,  we can derive the optimization as follows,
\begin{equation}
\begin{aligned}
\tW^*
 &= \argmax_{\tW} \sum_{u, v, r} \y_{u}^\top\B_{r}\y_{v} \mathbf{\uW}_{uvr} -\beta \|\tW\|_1\\
 &= \argmax_{\tW}\sum_{r, p,q}\mB_r[p, q] \sum_{\{u,v\}: (y(u)=p) \wedge (y(v)=q)}  \tW_{uvr}- \beta \|\tW\|_1
 \end{aligned}
\end{equation}
where $\mB_{r}[p,q]=P(w_{uvr}=1\mid \vy_{u,p} = \vy_{v, q}=1)$ with $\vy_{u,p}$ the p-th entry of the vector $\vy_u$ and $y_u$ is the label index of node $u$. The sum is over two parts: 1) over all relation types within one node type combination (in many scenarios this is only one) and 2) over all possible node labels combination. 

Then we define a ``representative'' connection between two nodes within  a combination of node types, that gives the highest probability of connection $\mB_r [p^*, q^*]$, we can derive the summation as,
\begin{equation}
\begin{aligned}
&\sum_{r} \left\{(\mB_{r}[p^*,q^* ] -  \sum_{(p,q)\neq (p^*, q^*)} 
\mB_{r}[p,q])\sum_{(y_u, y_v) =(p^*, q^*)}\tW_{uvr}\right\} \\+ &\sum_r \left\{\sum_{(p,q) \neq(p^*, q^*)} \mB_{r}[p,q ] (\sum_{(y_u, y_v)=(p^*, q^*)} \tW_{uvr}
+  \sum_{(y_u, y_v)\neq (p^*, q^*)} \tW_{uvr})\right\}-\beta\|\tW\|_1\\
&=\sum_{r}\left\{\underbrace{(\mB_{r}[p^*,q^*] -  \sum_{(p, q)\neq (p^*, q^*)} \mB_{r}[p,q ])\sum_{(y_u, y_v) = (p^*, q^*)}\tW_{uvr}}_{\text{Homophily Term}}\right\} + \underbrace{(\sum_r \sum_{(p, q) \neq (p^*, q^*)} \mB_{r}[p,q]-\beta ) \|\tW\|_1 }_{L^1 \text{ Regularizer}}\\
\end{aligned}
\end{equation}
The algorithm aims at maximizing the above objective. The first part is to assign larger weights on the edges connecting homophily nodes, and the second part is a $L^1$ regularizer enforcing the sparsity on $\tW$. The second regularizer has to have a negative coefficient and the first term need to have a positive coefficient. Thus, the sufficient (not necessary) condition for the optimization problem to have meaningful solutions is stated as follows. 

\textbf{Optimization Condition: } $\forall r \in \gR$, there exists a pair $(p^*, q^*)$ that satisfies,
\begin{equation}
\mB_{r}[p^*,q^* ] -  \sum_{(p,q)\neq (p^*, q^*)} 
\mB_{r}[p,q] >0
\end{equation}

\paragraph{Homophily Ratio on Heterogeneous graphs.} From the above optimization condition, the homophily ratio is consequently defined as,
\begin{equation}
\text{HR}(\G, r) = \frac{\mB_{r}[p^*,q^* ]}{\sum_{(p,q)\neq (p^*, q^*)} 
\mB_{r}[p,q])}
\end{equation}

We then show that the homophily ratio is linked to the relaxed homophily ratio defined in~\cref{eq: relaxed_hr}. 
\begin{equation}
\text{RHR}\left(\mathcal{G}, \Phi\right)=\frac{\sum_{\{u, v\} \in \mathcal{E}_{\Phi}} \mathbb{I}\left(y_u=y_v\right)}{\left|\mathcal{E}_{\Phi}\right|},
\end{equation}
\begin{propensity}
    The HR and RHR are positively related.
\end{propensity}
\textit{Proof.} Consider a meta-path $\Phi$ consisting of relations $R_1, R_2, \ldots, R_{L-1}$.

We first compute the probability of having two nodes, $u$ and $v$ connected by a path, share the same label, i.e.,  $y_u=y_v = p$. We denote two nodes connected via a meta-path $\Phi$ as $u \leftrightarrow v \text{ via } \Phi$. So that, $P\left(u \leftrightarrow v\right.$ via $\left.\Phi \mid y_u=y_v=p_0\right)$ is:

$$
P_{\Phi}(p_0)= P\left(u \leftrightarrow v \text{ via } \Phi \mid y_u=y_v=p_0\right) = \sum_{p_1, \ldots, p_{L-1}}\prod_{l=0}^{L-1} \mB_{R_l}\left[p_l, p_{l-1}\right]
$$
where we assume the labels along the path are $p_0, p_1, \ldots, p_{L-1}$ and $p_L=p_0$.

Thus, the expected value of the relaxed homophily ratio is the expected ratio of same-label connections over all the possible paths. This is derived by the probability of having the starting node and targeting node with the same labels:
\begin{equation}
\begin{aligned}
\mathbb{E}(\text{RHR})&=\sum_{p} P\left(y_u=p\right) \cdot P\left(y_v=p\right) \cdot P\left(u \leftrightarrow v \text { via } \Phi \mid y_u=y_v=p\right)\\
&=\sum_{p} P\left(y_u=p\right) \cdot P\left(y_v=p\right) \cdot \sum_{p_1, \ldots, p_{L-1}}\prod_{l=0}^{L-1} \mB_{R_l}\left[p_l, p_{l-1}\right]
\end{aligned}
\end{equation}
We consider the marginal distribution $P(y_u=p)$ is a uniform distribution , we have
\begin{equation}
\text{RHR}\propto\sum_{p} \sum_{p_1, \ldots, p_{L-1}}\prod_{l=0}^{L-1} \mB_{R_l}\left[p_l, p_{l-1}\right]
\end{equation}
We focus on the dominant path, which is the path along the ``representative relationship'' $\mB_{R_l}\left[p_l^*, p_{l-1}^*\right]$ for all $l$. This decomposes the RHR as,
$$\text{RHR}\propto\sum_{p} \left\{\underbrace{\prod_{l=0}^{L-1} \mB_{R_l}\left[p_l^*, p_{l-1}^*\right]}_{\text{Dominant Term}} + \underbrace{\sum_{p_1, \ldots, p_{L-1}\neq p_1^*\ldots p_{L-1}^*}\prod_{l=0}^{L-1} \mB_{R_l}\left[p_l, p_{l-1}\right]}_{\text{Other Terms}}\right\}$$
If we assume $\sum_{p,q}\mB_r[p, q] = \text{Constant}$ and $\mB_r[p^*, q^*] \geq \sum_{(p,q) \neq (p^*, q^*)}\mB_r[p, q] $ for each $r$, the RHR increases if any element in the dominant term, $\mB_{R_l}[p_l^*, p_{l-1}^*]$, increases. It is easy to see that the HR is positively related to $\mB_{r}[p^*, q^*]$. So this can show that the homophily ratio (HR) we defined, is positively related to the relaxed homophily ratio (RHR)~\citep{DBLP:conf/www/GuoDBFMCH0Z23}.

\section{ADDITIONAL EXPERIMENTAL RESULTS}

\subsection{The dataset statistics}

The dataset statistics are shown in~\cref{tab: lpresult}. With $|\gV|$ and $|\gE|$ the number of nodes and edges, $\gR$ the relation types, HR the homophily ratio and SDOR the Smoothest-dimension Overlapping ratio.  

\begin{table*}[ht]
\renewcommand\arraystretch{1}
\center
\begin{tabular}{llcccc}
\toprule 
Dataset &$|\mathcal{V}|$&$|\mathcal{E}|$  & $\mathcal{R}$ &  RHR  &SDOR\\
\hline 
\multirow{2}{*}{IMDB} &\multirow{2}{*}{11k}&\multirow{2}{*}{550k}& Movie-Actor (MA) & 0.510  & \multirow{2}{*}{MA-MD: 0.65} \\
&&& Movie-Director (MD) & 0.900  \\
\hline

\multirow{3}{*}{ACM} &\multirow{3}{*}{21k}&\multirow{3}{*}{87k} &Paper-cite-Paper (PP)& 0.64 & PP-PA: 0.66 \\ 
&&& Author-write-Paper (PA) & 0.925   & PP-PS: 0.67 \\
&&& Paper-has-subject (PS) & 0.804  & PA-PS: 0.94\\
\hline
\multirow{3}{*}{Yahoo} &\multirow{3}{*}{100}&\multirow{3}{*}{/} &Finance-Health& / & / \\ 
&&& Finance-Tech & /  & / \\
&&& Tech-Health& /& /\\
\bottomrule
\end{tabular}
\caption{Real-world datasets statistics}
\label{tab: lpresult}
\end{table*}

\subsection{Additional Quantitative Results}

We conduct analysis studies according to the embedding size, the homophily ratio and the optimization methods and the results can be found in~\cref{tab: new_algorithm_result}. The results illustrated that:
\begin{enumerate}
    \item The algorithm is more robust when the homophily ratio is higher, as we discussed in~\cref{sec: rbst_analysis}.
    \item The iterative refining optimization is more stable in ensuring convergence and thus yields better performance, as we suggested in~\cref{sec: quan_results}.
\end{enumerate}

\begin{table}[H]
\centering
\resizebox{\textwidth}{!}{\begin{threeparttable}
\renewcommand\arraystretch{0.6}
\center
\begin{tabular}{llcccccccccc}
\toprule 
& \multirow{2}{*}{\bf Dataset} & \multirow{2}{*}{\bf RHR} & \multirow{2}{*}{\bf K}
& \multicolumn{2}{c}{\bf Vanilla GSL} & \multicolumn{2}{c}{\bf HGSL-GD} & \multicolumn{2}{c}{\bf HGSL-IR} \\
\cmidrule(r){5-6} \cmidrule(r){7-8} \cmidrule(r){9-10} 
&&&& AUC & GMSE & AUC & GMSE & AUC & GMSE  \\

\midrule
&\multirow{3}{*}{Synthetic}  
&\multirow{3}{*}{0.95+}

&60
&0.66 $\pm$ 0.05& 0.03 $\pm$ 0.00 
&0.70 $\pm$ 0.04&0.27 $\pm$ 0.05
&\textbf{0.75 $\pm$ 0.02}&\textbf{0.02 $\pm$ 0.00} \\

&&&300
&0.65 $\pm$ 0.01&0.03 $\pm$ 0.00
&0.75 $\pm$ 0.05 &0.30 $\pm$ 0.04
&\textbf{0.83 $\pm$ 0.04}&\textbf{0.02 $\pm$ 0.01} \\
&&&600
&0.78 $\pm$ 0.03& \textbf{0.04 $\pm$ 0.02}
&0.77 $\pm$ 0.04 &0.26 $\pm$ 0.04
& \textbf{0.89 $\pm$ 0.04}&0.06 $\pm$ 0.01 \\

\midrule

&IMDB &0.51& 3066
&0.74 $\pm$ 0.04&0.29 $\pm$ 0.08
&0.75 $\pm$ 0.06&0.21 $\pm$ 0.05
&\textbf{0.81 $\pm$ 0.07}&\textbf{0.07 $\pm$ 0.05} \\

\midrule

&ACM & 0.64
&1902
&0.65 $\pm$ 0.06&\textbf{0.07$\pm$ 0.10}
&0.62 $\pm$ 0.06 &0.09 $\pm$ 0.15
&\textbf{0.73 $\pm$ 0.02}&0.12 $\pm$ 0.07\\

\bottomrule
\end{tabular}
\caption{Quantitative experimental results on Heterogeneous Graph Structure Learning.}
\label{tab: new_algorithm_result}
\begin{tablenotes}
\item $^*$ Experimental results are evaluated over 30 trials and the mean/standard deviation is calculated. 0.00 means the value $<0.01$.
\end{tablenotes}
\end{threeparttable}}
\end{table}

\vfill

\subsection{Analysis on the relation-wise embedding}

In synthetic dataset, we conduct the analysis of $\mE$ by measuring the normalized root mean squared error (NRMSE) of the estimated value and the ground truth by

$$
\frac{1}{|\mathcal{R}|} \sum_r \frac{\sqrt{\sum_k\left(\boldsymbol{e}_{r, k}-\hat{\boldsymbol{e}}_{r, k}\right)^2} / K}{\max \left(\boldsymbol{e}_r\right)-\min \left(\boldsymbol{e}_r\right)}
$$

where $r$ is the relation type, $k=[1, \ldots, K]$ is the entry on each signal dimension. We report the results below:
\begin{table}[H]
\centering
\begin{tabular}{llll} 
\toprule
Feature Dimension & K = 60 & K = 100 & K=300 \\
\hline NRMSE & $0.0320 \pm 0.0006$ & $0.0189 \pm 0.0003$ & $0.0063 \pm 0.0001$ \\
\bottomrule
\end{tabular}
\end{table}

The error is indeed small, which demonstrates the efficacy of our algorithm in learning a proper $\ve_r$. 

To better understand the property of $\ve_r$, we can set $\alpha=\frac{\lambda_2}{2 \lambda_1^2}$ and $\beta=\frac{\lambda_2}{\lambda_1}$ and rewrite~\cref{eq: FR} as,

$$
\ve_{r, k}^{t+1}=\frac{\lambda_2}{2 \lambda_1}\left(\frac{1}{\lambda_1} \sum_{v, u, r \in \gE_r^{\prime}} w_{v u r}^t \vx_{v, k} \cdot \vx_{u, k}-2\right)=\alpha \sum_{v, u, r \in \gE_r^{\prime}} w_{v u r}^t \vx_{v, k} \cdot \vx_{u, k}-\beta
$$

An intuition for this solution, as illustrated in ~\cref{fig: gen_smooth}, is that: If we consider $\beta \rightarrow 0$ and the features for each node $\mX_v$ is normalized to 1 , the solution of $\ve_r$ will put larger weights on the dimensions that exhibit higher similarity, and those dimensions are considered to be most informative in recovering the ground truth $\tW$ under our assumption. This phenomenon is observed in synthetic data, where the algorithm puts larger weights on the dimensions exhibiting higher smoothness.

\end{document}